%% file: combined.tex
\documentclass[runningheads]{llncs}

\usepackage[mobile]{eccv}

\usepackage{eccvabbrv}
\usepackage{tcolorbox}
\tcbuselibrary{breakable}
\usepackage{algorithmicx}
\usepackage{algorithm}
\usepackage[noend]{algpseudocode}
\usepackage{graphicx}
\usepackage{booktabs}
\usepackage{multirow}
\usepackage{microtype}
\usepackage{subfig}
\usepackage{float}
\usepackage{placeins}
\usepackage{enumitem}
\setlist[enumerate,2]{label=\arabic*.}
\usepackage[accsupp]{axessibility}
\usepackage[page]{appendix}

\usepackage[pagebackref,hypertexnames=false]{hyperref}
\usepackage{orcidlink}

\newcommand{\ours}{ChEaP}
\newcommand{\appref}[1]{Supp.~\ref{#1}}
\newcommand{\para}[1]{\smallskip\noindent\textbf{#1}}
\newcommand{\tablestyle}[2]{\setlength{\tabcolsep}{#1}\renewcommand{\arraystretch}{#2}\centering\footnotesize}
\definecolor{forestgreen}{HTML}{2E6F40}

\begin{document}

\input{main_content}

\bibliographystyle{splncs04}
\bibliography{main, ChEaP}


\appendix

\renewcommand{\thetable}{\Alph{section}\arabic{table}}
\renewcommand{\thefigure}{\Alph{section}\arabic{figure}}
\counterwithin{table}{section}
\counterwithin{figure}{section}

\newpage
\title{Supplementary Material}
\author{}
\date{}
\institute{}
\maketitle
\newcommand{\tocsec}[2]{\noindent\textbf{\hyperref[#1]{#2}}\dotfill\pageref{#1}\par}
\newcommand{\tocsubsec}[2]{\noindent\hspace{1.5em}\hyperref[#1]{#2}\dotfill\pageref{#1}\par}

\noindent\textbf{Table of Contents}
\smallskip

\tocsec{sec:impl-details}{A\;\, Implementation Details}
\tocsubsec{sec:traj-extraction}{A.1\quad Trajectory Extraction Pipeline}
\tocsubsec{sec:success-criteria}{A.2\quad Success Criteria}
\tocsubsec{sec:verifier-details}{A.3\quad Verifier Details}
\tocsubsec{sec:gen-hyperparams}{A.4\quad Generation Hyperparameters}
\tocsubsec{sec:text-prompts}{A.5\quad Text Prompts}

\tocsec{sec:sensitivity}{B\;\, EPBS Sensitivity Analysis}
\tocsubsec{sec:ablation-tau}{B.1\quad Ablation on Probe Step $\tau$}
\tocsubsec{sec:ablation-K}{B.2\quad Ablation on Beam Size $K$}
\tocsubsec{sec:wallclock}{B.3\quad Wall-Clock Comparison}

\tocsec{sec:extended}{C\;\, Extended Analysis}
\tocsubsec{sec:crossmodel-lockin}{C.1\quad Cross-Model Early Plan Commitment}
\tocsubsec{sec:hunyuan-deep-dive}{C.2\quad HunyuanVideo-1.5 Analysis}
\tocsubsec{sec:diagnostic-mazes}{C.3\quad Diagnostic Maze Variants}
\tocsubsec{subsec:supp_trajectory_diversity}{C.4\quad Trajectory Diversity Across Seeds}

\tocsec{sec:qualitative}{D\;\, Additional Qualitative Examples}
\tocsubsec{sec:gallery-commitment}{D.1\quad Early Commitment Gallery}
\tocsubsec{sec:gallery-chaining}{D.2\quad Chaining Gallery}
\tocsubsec{sec:gallery-failures}{D.3\quad Failure Mode Gallery}

\input{supp_content}

\end{document}

%% file: main_content.tex
\title{Video Models Reason Early: Exploiting Plan Commitment for Maze Solving}

\author{Kaleb Newman \and
Tyler Zhu \and
Olga Russakovsky}

\authorrunning{K.~Newman et al.}

\institute{Princeton University}

\maketitle

\begin{abstract}
Video diffusion models exhibit emergent reasoning capabilities like solving mazes and puzzles, yet little is understood about \emph{how} they reason during generation.
We take a first step towards understanding this and study the internal planning dynamics of video models using 2D maze-solving as a controlled testbed.
Our investigations reveal two findings. 
Our first finding is \textbf{early plan commitment}: video diffusion models commit to a high-level motion plan within the first few denoising steps, after which further denoising alters visual details but not the underlying trajectory.
Our second finding is that \textbf{path length}, not obstacle density, is the dominant predictor of maze difficulty, with a sharp failure threshold at 12 steps.
This means video models can only reason over long mazes by \textbf{chaining} together multiple sequential generations.
To demonstrate the practical benefits of our findings, we introduce \textbf{Ch}aining with \textbf{Ea}rly \textbf{P}lanning, or \ours{}, which only spends compute on seeds with promising early plans and chains them together to tackle complex mazes.
This improves accuracy from 7\% to 67\% on long-horizon mazes and by $2.5\times$ overall on hard tasks in Frozen Lake and VR-Bench across Wan2.2-14B and HunyuanVideo-1.5.
Our analysis reveals that current video models possess deeper reasoning capabilities than previously recognized, which can be elicited more reliably with better inference-time scaling.

  \keywords{Video Models \and Reasoning \and Test-Time Inference}
\end{abstract}

\section{Introduction}
\label{sec:intro}

\begin{figure}[!t]
    \centering
    \includegraphics[width=0.99\textwidth]{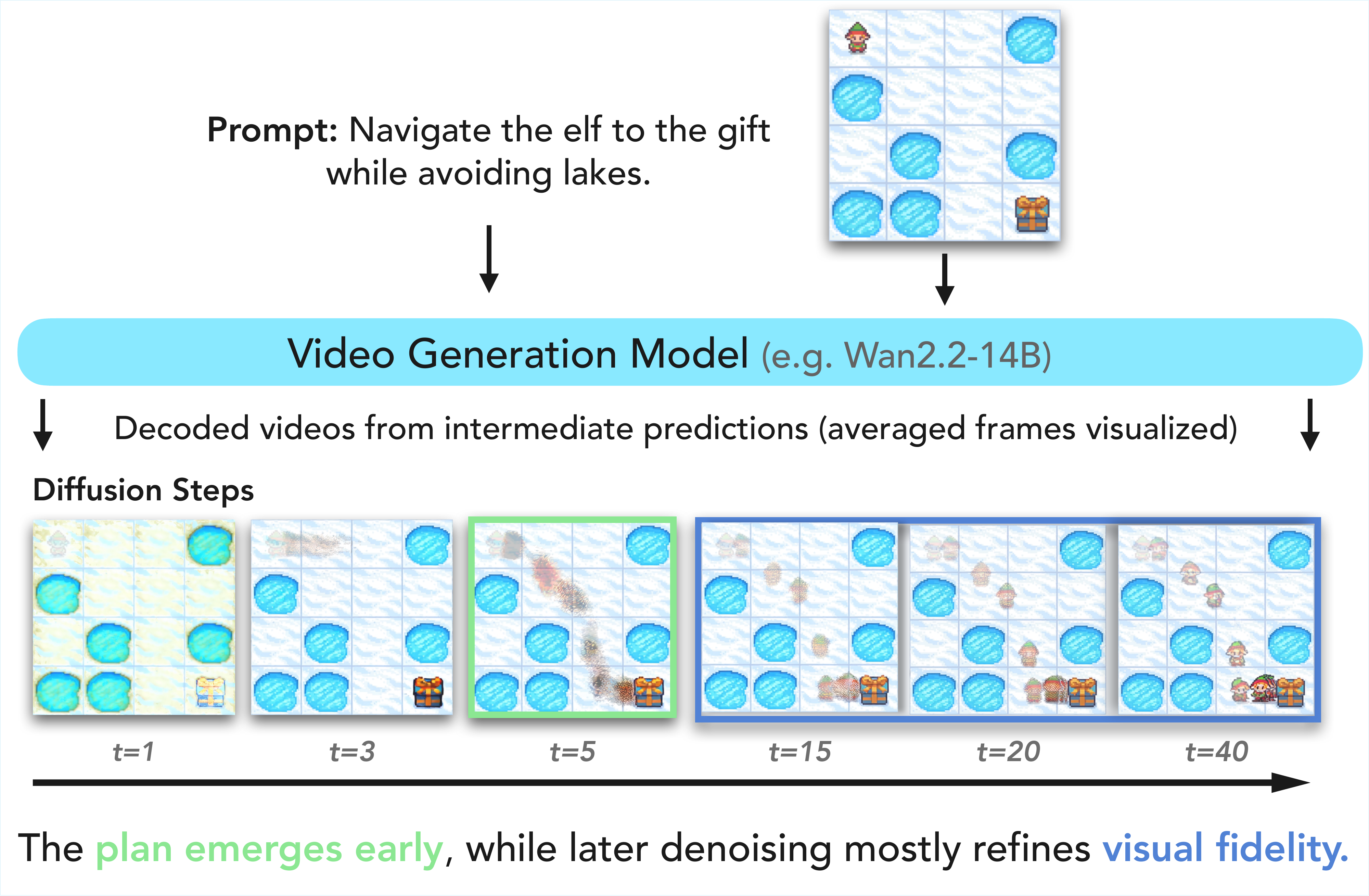}
    \caption{\textbf{Video diffusion models plan early.} 
Decoded intermediate $\hat{x}_0$ predictions reveal that the model commits to a trajectory within the first few denoising steps \textcolor{green!60!black}{(green box)}; later steps refine visual details but rarely alter the path \textcolor{blue}{(blue box)}. 
}
    \label{fig:teaser}
\end{figure}

Video generation models generate high-fidelity, temporally coherent videos that capture complex motion dynamics for creative applications or demonstrate intuitive world physics as synthetic data engines~\cite{wan_wan_2025,wu_hunyuanvideo_2025,agarwal2025cosmos}.
Recent works have discovered that these models exhibit emergent \emph{general-purpose vision understanding}, from basic perception and manipulation to maze and symmetry solving~\cite{wiedemer_video_2025,yang_reasoning_2025}.
Unlike vision-language models (VLMs), which map visual inputs to linguistic space, video models simulate reasoning directly in pixel space.
This makes video models particularly fit for spatial tasks like maze solving, object tracking, and robot navigating, which require a type of spatial imagination that some refer to as \emph{chain-of-frames} reasoning, i.e., using the frames as a visual scratchpad~\cite{wiedemer_video_2025}.
Yet despite growing interest in these capabilities, we lack a basic understanding of how such chain-of-frames reasoning \emph{emerges} during generation and how reliably we can elicit these latent capabilities for solving reasoning tasks.

Studying reasoning in open-ended video tasks, however, is challenging. 
Video models fabricate arbitrary details to produce diverse outputs, making them difficult to control.
Without a defined end goal, the outputs are even harder to verify automatically.
A controlled setting is much more tractable.

Maze solving provides exactly this.
Mazes have served as a canonical testbed for studying planning since Tolman's cognitive map theory~\cite{tolman1948cognitive}, through model-based reinforcement learning (RL)~\cite{sutton1991dyna}, to modern deep RL benchmarks~\cite{mirowski2016learning, chevalier2023minigrid}.
They require sequential, constraint-satisfying action planning, and conditioning on an input frame holds the environment constant, separating reasoning failures from rendering failures.
Ground-truth solutions exist via BFS, enabling automatic verification, and difficulty can be systematically varied through grid size, path length, and obstacle placement.
Using mazes as a controlled testbed, we analyze the internal dynamics of video diffusion models during maze solving and uncover a phenomenon we call \textbf{early plan commitment} (\cref{fig:teaser}).
This means the model commits to a high-level motion plan within the first few denoising steps, which remains stable for the remainder of sampling.

\begin{figure}[t]
    \includegraphics[width=0.99\textwidth]{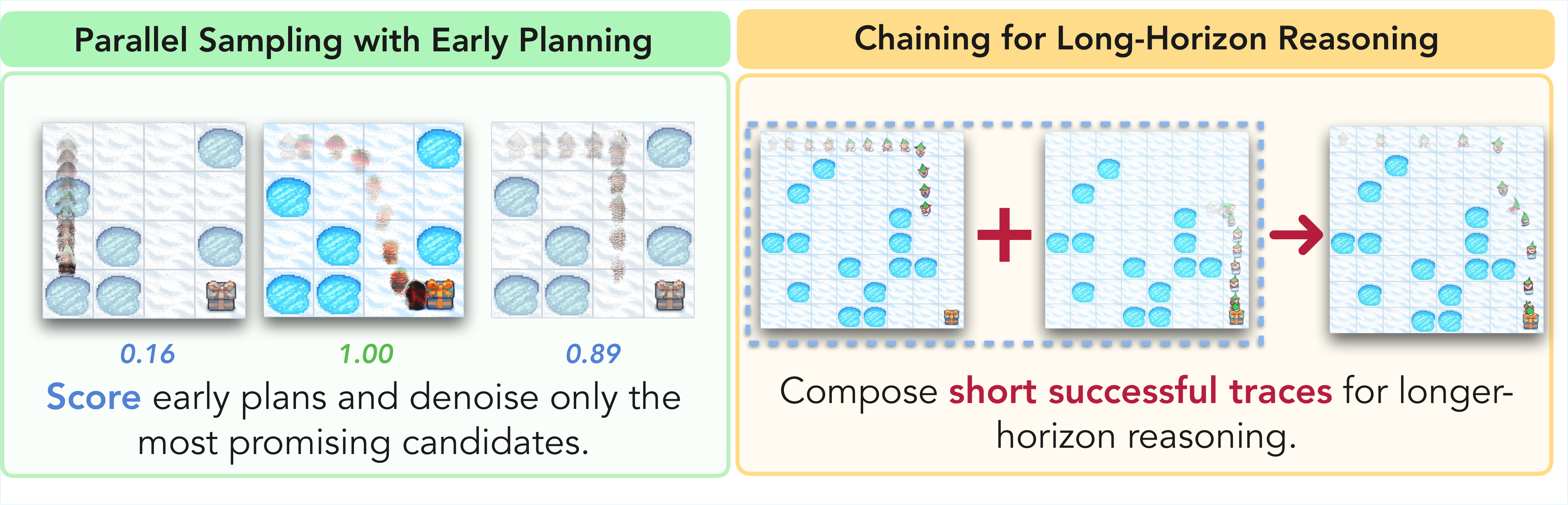}
    \caption{\textbf{Overview of ChEaP.} (Left) \emph{Early Planning Beam Search} scores early plans from partially denoised predictions and selects the most promising candidates for full generation. 
    (Right) \emph{Chaining} reconditions on the last frame of successful traces to extend reasoning beyond the single-generation horizon.}
    \label{fig:dual_method}
\end{figure}

This observation has immediate practical consequences.
If the plan is visible early, then standard best-of-$N$ sampling---which fully denoises every seed---wastes most of its compute polishing unsuccessful trajectories.
Instead, compute should be spent \emph{exploring more candidate plans} rather than refining each one.
We apply this insight with \emph{Early Planning Beam Search} (EPBS), which partially denoises a large pool of seeds, scores their early plans with a lightweight verifier, and reserves full denoising for only the most promising candidates.
We find this strategy is successful up until a sharp failure cliff at 12-step trajectories, which are too long for video models to solve in a single video. 
This further motivates \emph{chaining}: decomposing long-horizon tasks into shorter sub-problems to solve sequentially.
Together, we call this \textbf{Ch}aining with \textbf{Ea}rly \textbf{P}lanning, or ChEaP (\cref{fig:dual_method}).

Evaluated on the Frozen Lake~\cite{towers_gymnasium_2025} and VR-Bench~\cite{yang_reasoning_2025} datasets across the Wan2.2-14B~\cite{wan_wan_2025} and HunyuanVideo-1.5~\cite{wu_hunyuanvideo_2025} video models, \ours{} matches best-of-$N$ accuracy in $0.3\times$ the diffusion steps, achieves up to $2.5\times$ accuracy gains on hard tasks, and boosts long-horizon maze accuracy from 7\% to 67\%. 
Our analysis across two state-of-the-art models and over 480 mazes demonstrates the efficacy of ChEaP for enhancing maze solving capabilities, and that existing video models possess substantially deeper reasoning abilities than previously recognized.
Our project page is at \url{video-maze-reasoning.github.io}.

\section{Related Work}

\para{Visual reasoning in video models.}
Video diffusion models have recently demonstrated surprising emergent capabilities, solving mazes, puzzles, and physical reasoning tasks without task-specific training~\cite{wiedemer_video_2025}.
This has spurred new benchmarks and datasets for systematic evaluation~\cite{yang_reasoning_2025, wang_very_2026,chen_babyvision_2026}, as well as calls for process-aware metrics that go beyond final-frame accuracy~\cite{li_beyond_2026}.
Yet zero-shot generation still fails under strict long-horizon constraints~\cite{vo_vision_2025}.
He~\etal~\cite{he_diffthinker_2025} address this by fine-tuning the model for native image-to-image reasoning, but this requires retraining.
Rather than training the model, we show that stronger reasoning already exists within off-the-shelf video models and can be elicited through better \emph{inference-time} compute allocation.

\para{Phase transitions in diffusion models.}
Understanding \emph{why} inference-time strategies work requires examining the internal dynamics of diffusion.
It is now well established that the reverse diffusion process exhibits a coarse-to-fine hierarchy: global semantic structure crystallizes in the earliest denoising steps, while later steps refine only low-level detail~\cite{choi_perception_2022, balaji_ediffi_2022, raya_spontaneous_2023, sclocchi_phase_2024, biroli_dynamical_2024}.
Empirically, cross-attention maps~\cite{hertz_prompt--prompt_2023} and internal activations~\cite{kwon_diffusion_2023} confirm that spatial layout and semantic identity are fixed early, and stage-specialized denoisers exploit this by training separate experts per noise level~\cite{balaji_ediffi_2022}.
Theoretically, this behavior has been formalized as sharp phase transitions~\cite{raya_spontaneous_2023, biroli_dynamical_2024} with bounded critical windows~\cite{li_critical_2024}, and studied through geometric~\cite{yaguchi_geometry_2024}, spectral~\cite{ventura_manifolds_2025}, and information-theoretic~\cite{ramachandran_cross-fluctuation_2025} lenses.
However, these analyses focus on image generation.
We show that the same early-commitment behavior holds for \emph{video} diffusion, and that it applies not just to appearance, but to the model's motion plan.

\para{Inference-time scaling for diffusion models.}
Given this internal structure, a natural question is how to exploit it training-free.
Inspired by the success of test-time compute scaling in language models~\cite{snell_scaling_2024}, a growing line of work applies similar ideas to diffusion: searching over noise seeds with verifier feedback~\cite{ma_inference-time_2025, kim_inference-time_2025}, Feynman-Kac particle resampling~\cite{singhal_general_2025}, tree search with MCMC refinement~\cite{zhang_inference-time_2025}, and noise-trajectory optimization for visual quality~\cite{liu2025video}.
These methods improve output quality by generating and evaluating more candidates, but they treat the diffusion process as a black box, allocating compute uniformly across timesteps without exploiting \emph{when} the model commits to its plan.
Our EPBS exploits early plan commitment to prune unpromising seeds after only a few denoising steps, as we find that mid/late-stage refinement contributes little to reasoning diversity.

\section{Mazes as a Controlled Testbed for Studying Reasoning}
\label{sec:mazes_testbed}

We focus on mazes as a controlled proxy for \emph{action planning}---a setting that is verifiable, visually grounded, and sufficiently challenging to require multi-step decision-making.
Crucially, conditioning on an input frame fixes the maze layout, so all task-relevant structure is concentrated in the agent's \emph{motion trajectory}---making failures diagnostic and solutions automatically verifiable.

\para{Research questions.} Mazes are well positioned for studying the \emph{internal dynamics} of generation, because trajectories can be extracted from \emph{intermediate} denoising predictions and compared against exact solution structure.
Recent work has established \emph{that} video diffusion models can solve mazes~\cite{wiedemer_video_2025, yang_reasoning_2025}, and that sampling more candidates improves reliability by 10--20\%~\cite{yang_reasoning_2025}, but these results describe only the \emph{outputs} of the process.
We do not know when the model commits to a plan during denoising, what structural properties of a maze make it difficult, or why brute-force sampling plateaus on hard instances.
We ask:
\begin{enumerate}
\item \textbf{When does the model decide its answer when tasked with a reasoning problem?}
\item \textbf{Are there any signals early in the denoising process that are predictive of success?}
\item \textbf{What are the common failure modes and what structural properties of the task drive them?}
\end{enumerate}

\para{Experimental setup.} To answer these questions with precision, we need control over our dataset.
We evaluate on \textbf{Frozen Lake} mazes~\cite{towers_gymnasium_2025}, which has an elf in the top left cell whose goal is to reach a gift in the bottom right while avoiding falling into frozen lakes along the way.
We vary grid size ($4{\times}4$ to $10{\times}10$), obstacle density ($20$--$80\%$), and goal placement---distinguishing \textbf{norm} mazes (goal at the far corner, maximizing path length) from \textbf{vary} mazes (randomly placed goal, often admitting shorter solutions).
We complement this with \textbf{VR-Bench}~\cite{yang_reasoning_2025}, which tests whether the same planning behaviors hold across different visual textures and constraint types (maze navigation and trap avoidance).
Across these two benchmarks, we evaluate over $480$ maze environments on two state-of-the-art video diffusion models: Wan2.2-14B~\cite{wan_wan_2025} and HunyuanVideo-1.5~\cite{wu_hunyuanvideo_2025}, using the standard image+text-to-video paradigm\cite{blattmann2023stable, esser2023structure}. 
We provide examples in~\appref{sec:qualitative}.

\para{Evaluation.} We evaluate \emph{task success} rather than exact path match.
Exact-match evaluation, as used in prior VR-Bench work~\cite{yang_reasoning_2025}, is overly strict: many mazes admit multiple valid solutions, and small trajectory-extraction errors can invalidate an otherwise correct video.
Instead, we extract the agent trajectory from the generated video using SAM2~\cite{ravi_sam_2024} and mark a sample as successful if the agent reaches the goal without violating task constraints, while rejecting degenerate cases such as goal drift.
Full extraction details and success criteria are provided in \appref{sec:traj-extraction} and \appref{sec:success-criteria}.

\section{Early Plan Commitment in Video Diffusion Models}
\label{sec:early_trajectory}

Using mazes as a controlled testbed, we can analyze \emph{how} solutions form during denoising.
Because the answer to a maze is expressed primarily through the agent's motion path, we study how this trajectory evolves across intermediate $\hat{x}_0$ predictions. This analysis reveals a striking pattern: for the vast majority of seeds, the model commits to a coarse route within the first few denoising steps, and later computation primarily refines visual fidelity rather than changing the underlying plan. We refer to this phenomenon as \emph{\textbf{early plan commitment}}.

\subsection{Flow matching}
\label{subsec:flow_matching}

Both video models we study are built on \emph{flow matching}~\cite{lipman_flow_2023,liu_flow_2023,esser_scaling_2024}, a generative framework that learns a velocity field transporting noise $\epsilon$ to data $x_0$ along straight paths.
During training, an interpolant constructs noisy samples $x_t = (1-t)\,x_0 + t\,\epsilon$ for $t\in[0,1]$ and $\epsilon\sim\mathcal{N}(0,I)$, and a network $v_\theta$ is trained to predict the velocity $\frac{dx_t}{dt}=\epsilon - x_0$.
At inference, one integrates $v_\theta$ from $t=1$ (pure noise) to $t=0$ (clean video) using a discrete schedule of $T$ steps.

Crucially, at any intermediate step $t$ the velocity prediction can be rearranged to recover a \emph{clean-sample estimate}:
\begin{equation}
\label{eq:x0_pred}
\hat{x}_0^{(t)} = x_t - t \cdot v_\theta(x_t, t).
\end{equation}
This $\hat{x}_0$ prediction is the model's current best guess of the final video given only the partially denoised state.
Early in sampling, $\hat{x}_0^{(t)}$ is blurry and coarse, but as $t\to 0$ it converges to the final output.
We decode these intermediate predictions throughout denoising to study \emph{when} the model's plan takes shape.

\subsection{What is an early trajectory?}
\label{subsec:what_is_early_trajectory}

\begin{figure}[!t]
    \centering
    \includegraphics[width=0.98\textwidth]{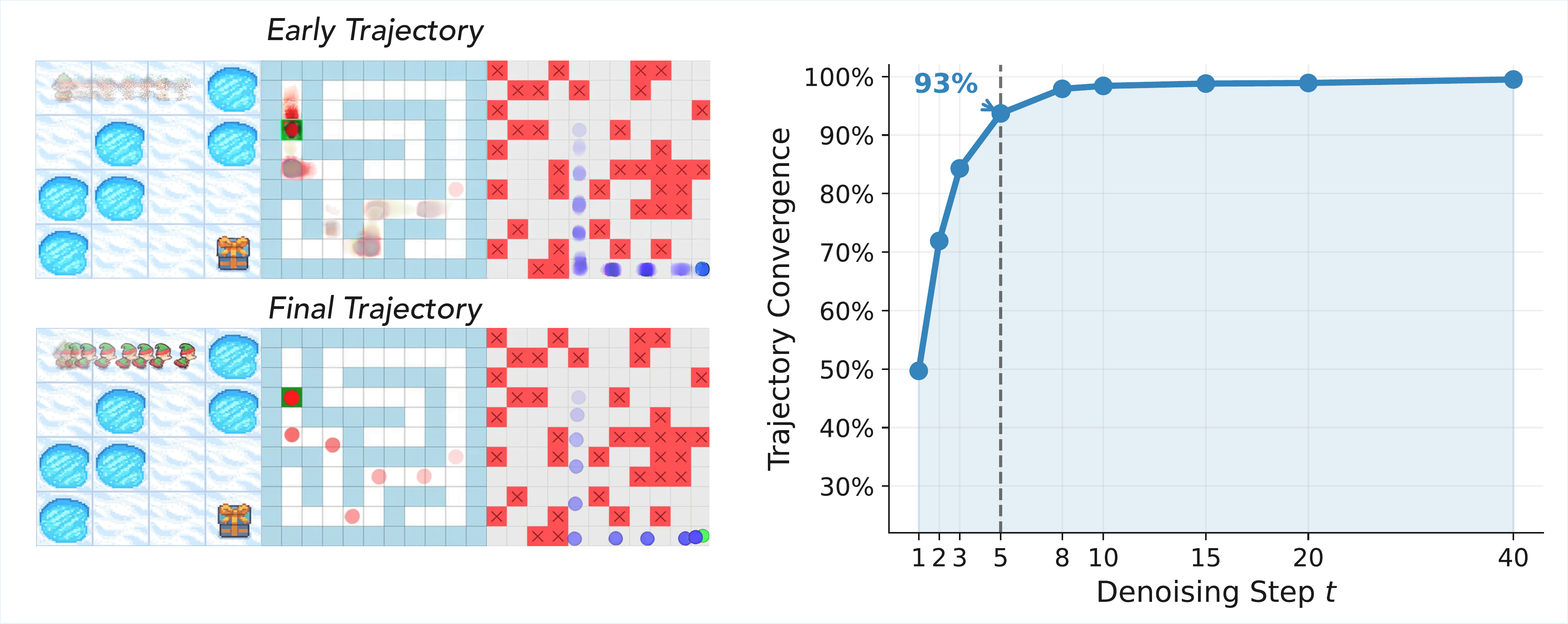}
    \caption{\textbf{Early plans stay consistent.} (Left) Across multiple settings, the early trajectories emerging from decoded $\hat{x}_0$ predictions at step 5 match the final trajectory. 
    (Right) Mean trajectory convergence throughout the denoising process. 
    Step 5 already reaches 93\%, \ie trajectories stay converged (over 163 $4{\times}4$ mazes).}
    \label{fig:plan_lockin}
\end{figure}

For a fixed random seed, let $\hat{x}_0^{(t)}$ denote the model's decoded prediction of the clean video at denoising step $t$ into pixel space.
We define an \emph{early trajectory} $\mathcal{T}^{(t)} = (c_0^{(t)}, c_1^{(t)}, \dots)$ as the sequence of cells this intermediate prediction visits.

This is the natural object to study in mazes because the trajectory carries almost all of the task-relevant structure. The maze layout, obstacles, and goal are meant to be fixed according to the conditioning image and prompt; only the agent moves. 
Thus, to understand planning in this setting, we do not need every intermediate prediction to be visually sharp or pixel-accurate. 
We need only ask whether the model has already committed to the \emph{route} it will ultimately follow.

\subsection{Trajectories converge early during denoising}
\label{subsec:trajectories_crystallize}

To quantify when trajectories converge, we calculate the \emph{trajectory convergence} $\mathcal{C}$ of a step $t$ prediction to the final one at time step $T$. 
We first extract a spatial \emph{motion energy map} from each video: for every grid cell, we count the total number of pixels whose color deviates from the estimated background across all frames (with the goal cell masked to suppress its idle animation). This yields an $N\times N$ matrix $\mathbf{M}^{(t)}$ summarizing where motion occurs, with high values along the elf's path and near-zero values elsewhere.
We then measure how well the intermediate energy pattern matches the final one via cosine similarity:
\[
\mathcal{C}(\text{step } t)
= \frac{\mathbf{m}^{(t)} \cdot \mathbf{m}^{(T)}}
       {\|\mathbf{m}^{(t)}\|\;\|\mathbf{m}^{(T)}\|},
\]
where $\mathbf{m}^{(t)}$ and $\mathbf{m}^{(T)}$ are the flattened energy vectors from the step-$t$ prediction and the final video, respectively. A convergence of $1.0$ indicates that motion energy is distributed identically across grid cells; $0.0$ indicates no agreement.
This metric is scale-invariant (robust to brightness differences between intermediate and final renderings) and requires no binarization threshold, avoiding the sensitivity to energy normalization that affects discrete cell-overlap measures on small grids (see supplement for a detailed comparison).

\Cref{fig:plan_lockin} shows that trajectory structure emerges surprisingly early.
On $4{\times}4$ Frozen Lake mazes, the step 5 trajectories for Wan2.2-14B are at $93\%$ mean convergence.
By step~10, convergence is nearly perfect. 
In other words, the model usually decides its route within the first quarter of denoising; the remaining steps primarily improve visual fidelity instead of altering the plan.

Because the metric operates on continuous energy values rather than binary cell sets, it is robust even on larger grids where background-difference extraction can introduce low-level noise in unvisited cells: such noise contributes zero to the numerator (since the reference energy is zero there) and only marginally affects the denominator.
We find the same pattern holds across sizes and models in \appref{sec:crossmodel-lockin}, and we include visual examples of early plan commitment in \appref{sec:gallery-commitment}.

\begin{figure}[t]
    \centering
    \includegraphics[width=\textwidth]{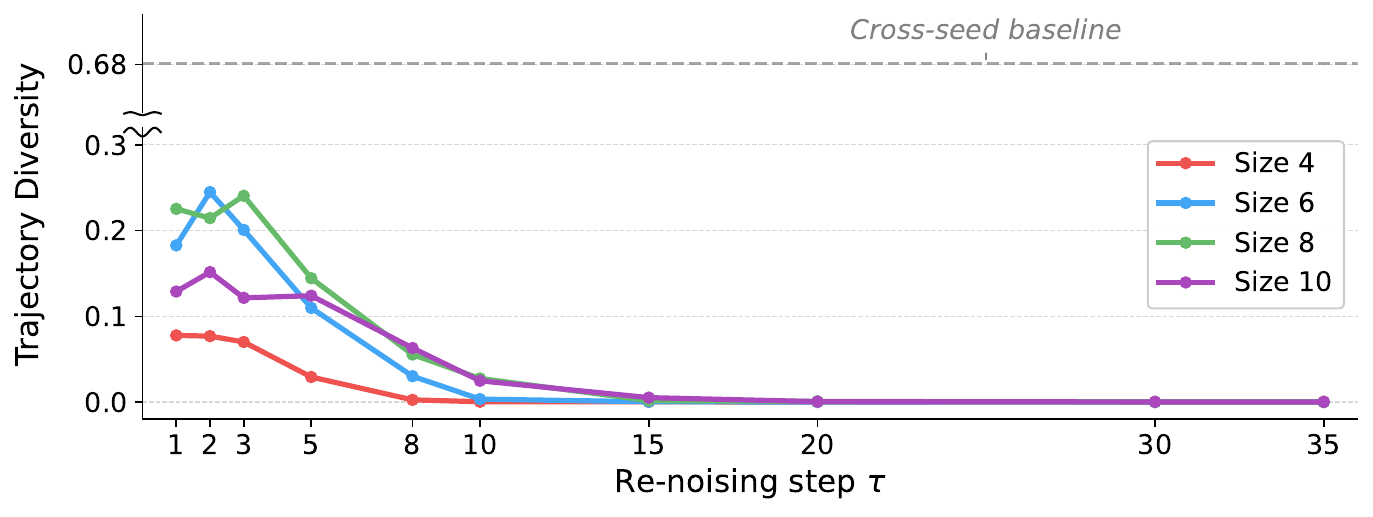}
    \caption{\textbf{Stepwise refinement.} Mean pairwise trajectory IoU among $K{=}5$ re-noised completions at each step~$\tau$, across grid sizes 4--10. Even at $\tau{=}1$, branch trajectories are far more similar to each other than trajectories from different seeds (dashed line), indicating that the route is largely encoded in the initial noise sample.
    }
    \label{fig:refinement_ablation}
\end{figure}

\subsection{Early trajectories are diverse across seeds, not refinement}
\label{subsec:why_early_trajectory_matters}

This observation suggests that reasoning-relevant structure is not uniformly distributed across denoising. 
A common technique for sample diversity in flow matching is \emph{refinement}, where the step $t$ noise is added back to $\hat{x}_0^{(t)}$ before continuing to denoise to explore other denoising paths.
We suspect that trajectory diversity benefits more from \emph{sampling different seeds}, not from refining an existing one. 
To test this, we perform a refinement ablation in which we renoise at a chosen step during the denoising process to sample a different trajectory (\cref{fig:refinement_ablation}).
We measure their diversity as $1 - \mathcal{C}(t', T)$ calculated between the new resulting trajectory from renoising at step $t$ and the old final trajectory at step $T$.

We find that refinement branches from the same seed are nearly identical in trajectory with at most 25\% trajectory diversity. 
Refining earlier in the denoising process or for larger mazes results in higher diversity, but none are as high as the 68\% diversity between different seeds.
\textbf{If early trajectories are predictive of final success, then inference-time scaling for reasoning should prioritize screening more candidate trajectories} rather than fully decoding every seed.

\section{Trajectory Screening for Efficient Sampling}
\label{sec:epbs}

The plan commitment phenomenon suggests a straightforward strategy: instead of fully denoising every seed, screen trajectories from early timesteps and discard unpromising ones. We formalize this as \emph{Early Planning Beam Search} (EPBS).

\subsection{Early Planning Beam Search}
\label{subsec:epbs}

EPBS reallocates inference-time compute from full denoising to early candidate exploration.
Rather than fully denoising every seed, we first partially denoise many candidates for $\tau$ steps, score their intermediate $\hat{x}_0$ predictions with a lightweight verifier, and reserve full decoding only for the top-$K$ seeds (\cref{alg:epbs}).
Under a fixed number of function evaluations (NFEs), this allows EPBS to explore substantially more candidate trajectories than standard best-of-$N$ sampling.

\begin{algorithm}[!t]
\begin{algorithmic}[1]
\Require video model with denoising steps $T$, budget $B$, probe step $\tau$, beam size $K$
\State Compute the number of initial candidates:
$N = \left\lfloor \frac{B-KT}{\tau} \right\rfloor + K$
\State Sample $N$ random seeds
\For{each seed}
    \State Partially denoise for $\tau$ steps
    \State Decode the intermediate $\hat{x}_0$ prediction
    \State Score the decoded prediction with the verifier
\EndFor
\State Select the top-$K$ seeds under the verifier score
\State Fully denoise only these $K$ candidates to $t=0$
\State Return the highest-scoring final sample
\end{algorithmic}
\caption{Early Planning Beam Search (EPBS)}
\label{alg:epbs}
\end{algorithm}

For Wan2.2-14B ($T=40$), EPBS with $\tau=5, K=1$ evaluates 73 candidates at $B=400$, compared to only 10 for best-of-$N$. 
The gain comes from terminating most seeds after only a small fraction of the denoising schedule, relying on planning commitment to ensure the $\tau$-step trajectory predicts the final output.
All that remains is to score the $\hat{x}_0$ predictions with a verifier.

\para{Lightweight trajectory verifier.}
The verifier requires only minimal privileged information: the locations of the agent, the goal, and obstacle cells (lakes, traps, or walls depending on the environment).
We argue that this is a reasonable setup; the goal is fully observable in a 2D setting, only the path to it must be discovered.

To score intermediate $\hat{x}_0$ predictions, we track the agent's position across frames and compute a confidence score that rewards goal progress while penalizing time spent in obstacle cells.
Only the top-$K$ seeds ranked by this score are fully decoded.
Full verifier details and the scoring formula are provided in \appref{sec:verifier-details}.

\subsection{Results}
\label{subsec:epbs_results}

We compare EPBS to best-of-$N$ sampling (equivalent to $\tau=T$) with $N=1,\dots,10$. We use beam size $K=2$ for both methods with the pass@$K$ metric~\cite{chen_evaluating_2021}, following prior work~\cite{wiedemer_video_2025}, and test on Wan2.2-14B ($T=40$) and HunyuanVideo-1.5 Step-Distilled ($T=8$).

\begin{figure}[!t]
    \centering
    \includegraphics[width=0.95\textwidth]{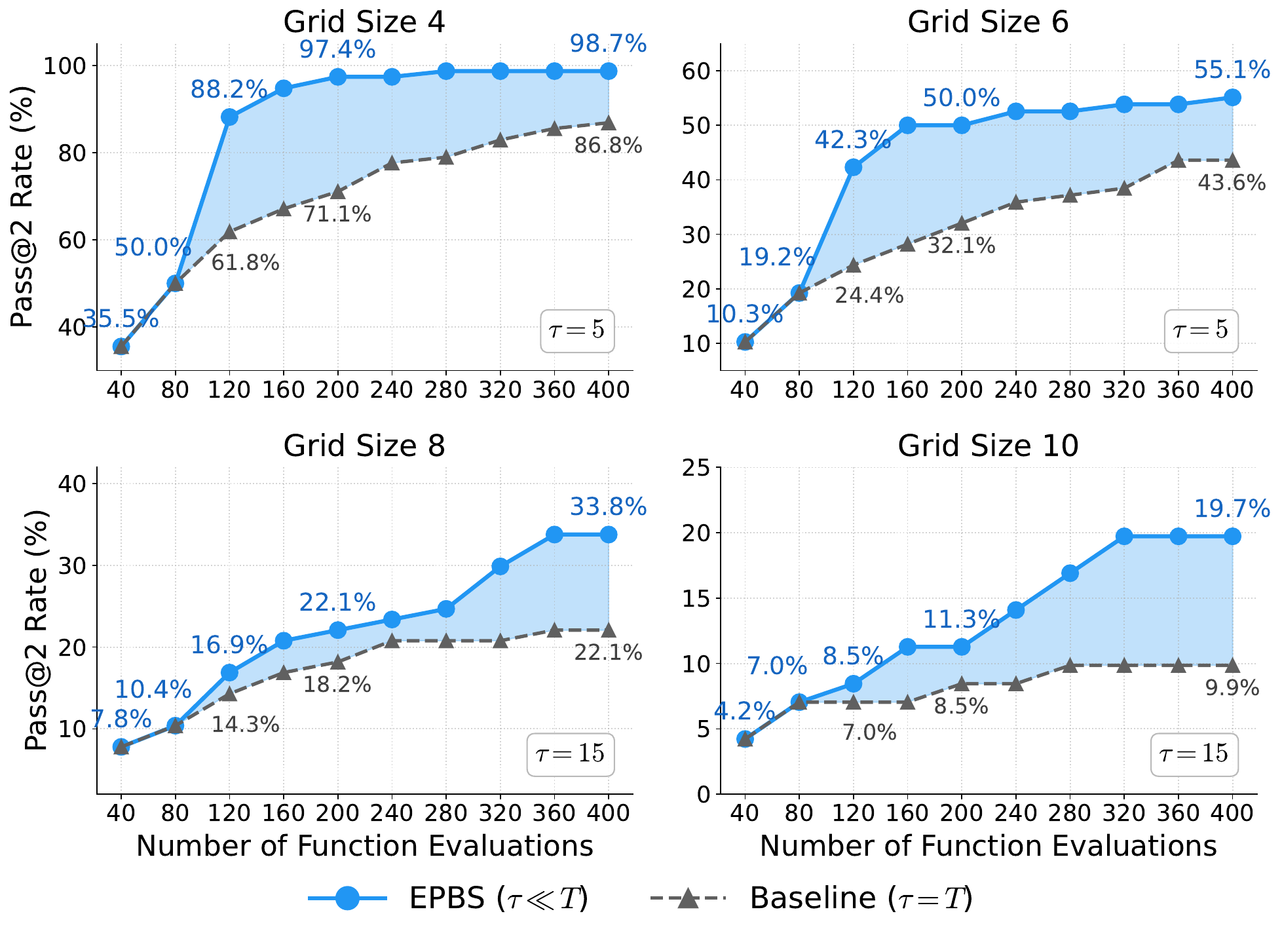}
    \caption{\textbf{EPBS finds solutions much more efficiently than best-of-$N$.} Accuracy vs Function Evaluations (NFEs) on Frozen Lake mazes across four sizes with Wan2.2-14B. EPBS consistently dominates standard best-of-$N$, with large gains on larger mazes.}
    \label{fig:acc_vs_nfe_all_sizes}
\end{figure}

As shown in \cref{fig:acc_vs_nfe_all_sizes}, EPBS consistently outperforms best-of-$N$ by ${\sim}10\%$ on average for Wan2.2-14B across all maze sizes. It matches best-of-$N$ accuracy with $3.3\times$ fewer NFEs and especially shines on large mazes (size 10), where exploring more candidate seeds breaks through plateaus reached by standard sampling. 
We see similar benefits on HunyuanVideo-1.5, with 13\% pass@2 on $4\times 4$ mazes and 3-4\% improvements for larger mazes and VR-Bench in \cref{tab:main_results}.

We acknowledge that NFE is not a complete cost measure, since each $\hat{x}_0$ probe requires a VAE decode (${\sim}1.5$ FEs in wall-clock). 
We provide wall-clock comparisons for completeness (\appref{sec:wallclock}), and find our takeaways still hold.

\para{Why EPBS works: early predictions are reliable.}
Our verifier reliably identifies promising seeds from early $\hat{x}_0$ predictions.
As shown in \Cref{tab:verifier_precision}, the verifier's top-2 selections succeed $2.2\times$ more often than random on easy mazes and $5.5\times$ on hard ones.
The ROC AUC of the verifier's confidence score against final success is above 0.85 across all sizes, confirming that the ranking is informative rather than merely noisy filtering.
To check whether the verifier discards otherwise correct solutions, we also compute an oracle score that returns success whenever \emph{any} seed in the pool solves the maze.
The gap between the verifier's top-2 accuracy and this oracle is at most 1.4\% on all sizes except~6, indicating that when a solution exists in the candidate pool, the verifier almost always finds it.

\para{Ablations.}
We fully ablate $\tau$ and $K$ on Frozen Lake mazes in Supp. \ref{sec:ablation-tau} and \ref{sec:ablation-K}, but summarize our results here.
Probing at $\tau=5$ yields the best trade-off for smaller sizes, while $\tau=10,15$ is better for larger mazes, indicating that trajectory convergence happens later.
Beam size $K=2$ is best at low budgets while $K=1$ works fine at higher budgets when the verifier becomes more reliable.

\section{Chaining generations for long-horizon reasoning}
\label{sec:chaining}

EPBS improves seed selection by exploiting early plan commitment, yet performance still collapses on large mazes.
Even an oracle that always picks the best seed from the candidate pool cannot succeed, because no single generation contains a complete solution. The bottleneck is structural.
Therefore, we ask: what structural properties make a maze hard, and where does the model's capability actually break down? 
\subsection{What makes a maze hard?}
\label{subsec:what_makes_a_maze_hard}

To understand where EPBS breaks down, we analyze which structural properties of a maze predict difficulty.

\begin{table*}[!t]
    \centering
    \begin{minipage}{.475\linewidth}
        \centering
        \captionof{table}{\textbf{Verifier reliability.}
        The verifier remains informative even on the hardest mazes, where successful seeds are rare.}
        \vspace{-0.7em}
        \resizebox{\textwidth}{!}{
        \tablestyle{4pt}{1.05}
        \begin{tabular}{lcccc}
            Size & $4\times 4$ & $6\times 6$ & $8\times 8$ & $10\times 10$ \\
            \specialrule{.15em}{.075em}{.075em}
            Random        & 40.7 & 14.9 & 3.9  & 1.8 \\
            Verifier top-$2$ & 90.7 & 41.0 & 21.4 & 9.9 \\
            \textcolor{gray}{Oracle} & \textcolor{gray}{92.0} & \textcolor{gray}{51.3} & \textcolor{gray}{21.4} & \textcolor{gray}{11.3} \\
            \hline
            Gain ($\times$) & 2.2 & 2.7 & 5.5 & 5.5 \\
            AUC           & 0.901 & 0.855 & 0.953 & 0.939 \\
            \hline
        \end{tabular}
        }
        \label{tab:verifier_precision}
    \end{minipage}%
    \hspace{0.8em}
    \begin{minipage}{.475\linewidth}
        \centering
        \captionof{table}{\textbf{Maze difficulty.}
        The success rate decrease is driven by trajectory length rather than obstacle density.}
        \vspace{-0.7em}
        \resizebox{\textwidth}{!}{
        \tablestyle{4pt}{1.05}
        \begin{tabular}{lcccc}
            Size & $4\times 4$ & $6\times 6$ & $8\times 8$ & $10\times 10$ \\
            \specialrule{.15em}{.075em}{.075em}
            Overall & 98.7 & 55.1 & 33.8 & 19.7 \\
            \hline
            Norm    & 100.0 & 27.5 & 7.5  & 0.0 \\
            Vary    & 97.2 & 84.2 & 62.2 & 41.2 \\
            \hline
            Path corr. & -0.01 & -0.60 & -0.81 & -0.79 \\
            Lake corr. & 0.03 & -0.01 & -0.05 & 0.05 \\
            \hline
        \end{tabular}
        }
        \label{tab:difficulty_overview}
    \end{minipage}%
\end{table*}

\para{Path length dominates difficulty.}
\Cref{tab:difficulty_overview} shows a stark gap between norm mazes (fixed far-corner goal, maximally long paths) and vary mazes (random goal, often shorter paths): on size~8, norm mazes achieve only 7.5\% versus 62.2\% for vary mazes.
The gap is explained entirely by path length: the Pearson correlation between ground-truth path length and EPBS success is $r = -0.81$ on size~8 and $r = -0.79$ on size~10.
Counter-intuitively, lake density has near-zero correlation with success ($|r| < 0.05$).
Avoiding obstacles is not what limits the model; planning long sequential trajectories is.

\begin{figure}[!t]
    \centering
    \includegraphics[width=0.95\textwidth]{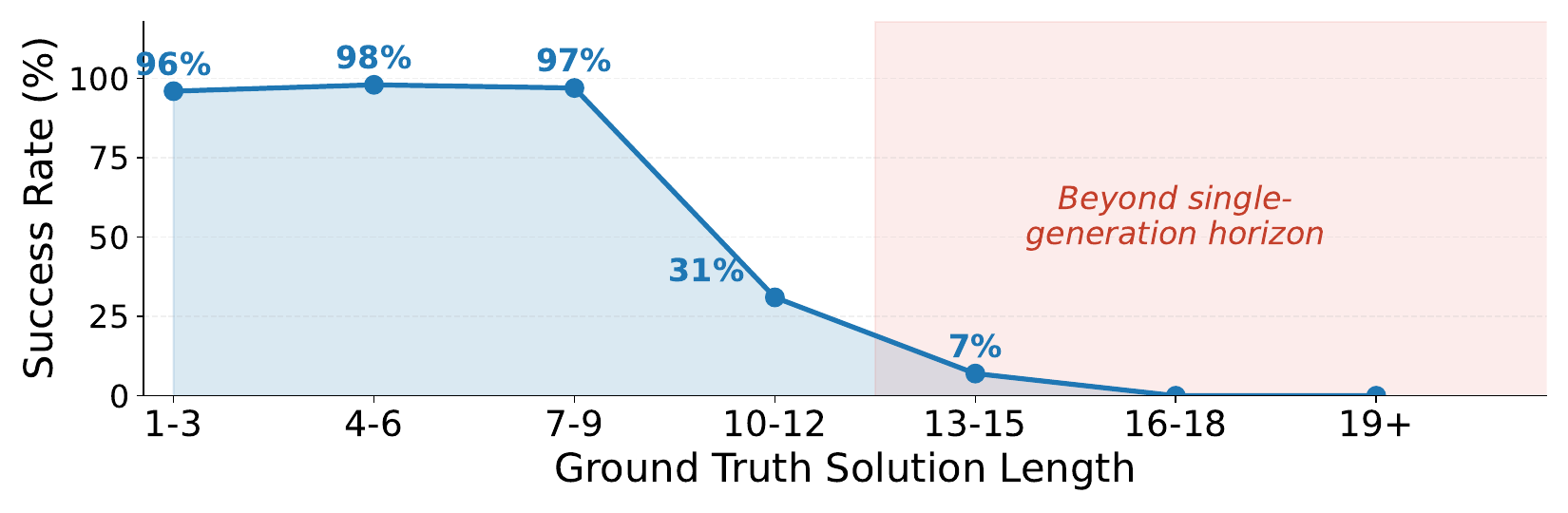}
    \caption{\textbf{EPBS fails beyond the single-generation horizon.}
While short paths are solved reliably, success drops sharply for trajectories longer than 10–12 steps. This breakdown persists even with strong seed selection, indicating a limitation in executing long plans rather than selecting them.}
    \label{fig:succ_v_path}
\end{figure}

\para{A sharp horizon threshold.}
Breaking this down by path length confirms the picture (\Cref{fig:succ_v_path}): the model reliably solves paths of ${\leq}\,9$ steps even on large grids, but drops below 10\% at 13 steps and beyond.
We hypothesize that the bottleneck is not seed selection or obstacle avoidance, but the model's \emph{generation horizon}. 
The video is too short for the agent to solve the entire maze. 
Since the model can plan short segments reliably regardless of grid size, a natural strategy is to decompose longer mazes into shorter segments and solve them sequentially.

\subsection{Chaining}
\label{subsec:chaining}

\begin{table}[!t]
\centering
\caption{\textbf{ChEaP substantially improves maze performance.}
We report pass@2 for best-of-$N$ (BoN), EPBS, and \ours{} (EPBS + Chaining) on Frozen Lake and VR-Bench.
For Wan2.2-14B, EPBS at 120 NFEs performs on par with BoN at 400 NFEs---a $3.3\times$ reduction in NFEs---and improves pass rate by 11.9 points on average.
}
\label{tab:main_results}
\setlength{\tabcolsep}{5pt}
\begin{tabular}{l c rrrr ccc}
\toprule
 & & \multicolumn{4}{c}{\textbf{Frozen Lake}} & \multicolumn{3}{c}{\textbf{VR-Bench}} \\
\cmidrule(lr){3-6} \cmidrule(lr){7-9}
\textbf{Method} & NFEs & $4\!\times\!4$ & $6\!\times\!6$ & $8\!\times\!8$ & $10\!\times\!10$ & Easy & Medium & Hard \\
\midrule
\multicolumn{9}{l}{\textit{Wan2.2-14B (40 NFEs / full gen)}} \\
    BoN  & 120 & \hfill 61.8 & \hfill 24.4 & \hfill 14.3 & \hfill 7.0  & \hfill 38.0 & 10.0 & 0.0 \\
    EPBS & 120 & \hfill 88.2 & \hfill 42.3 & \hfill 16.9 & \hfill 8.5  & \hfill 54.0 & 26.0 & 10.0 \\

    \addlinespace[4pt]
    BoN  & 400 & \hfill 86.8 & \hfill 43.6 & \hfill 22.1 & \hfill 9.9  & \hfill 56.0 & 28.0 & 10.0 \\
    EPBS & 400 & \hfill \textbf{98.7} & \hfill 55.1 & \hfill 33.8 & \hfill 19.7 & \hfill 68.0 & 44.0 & 25.0 \\
    ChEaP & 1200 & \hfill \textbf{98.7} & \hfill \textbf{88.5} & \hfill \textbf{46.8} & \hfill \textbf{22.5} & \textbf{72.0} & \textbf{48.0} & 25.0 \\
\midrule
\multicolumn{9}{l}{\textit{HunyuanVideo-1.5 Step Distilled (8 NFEs / full gen)}} \\
    BoN  & 40 & 27.6 & 14.1 & 11.7 & 1.4 & 28.8 & 2.8 & 3.7\\
    EPBS & 40 & 36.8 & 20.5 & 11.7 & 1.4 & 30.1 & 4.2 & 3.7 \\

    \addlinespace[4pt]
    BoN  & 80 & 38.2 & 20.5 & 11.7 & 1.4 & 30.1 & 2.8 & 3.7 \\
    EPBS & 80 & 51.3 & 24.4 & 14.3 & \textbf{5.6} & \textbf{35.6} & \textbf{4.2} & \textbf{3.7} \\
    ChEaP & 240 & \textbf{60.5} & \textbf{29.4} & \textbf{15.5} & \textbf{5.6} & \textbf{49.3} & \textbf{4.2} & \textbf{3.7} \\
\bottomrule
\end{tabular}
\vspace{-2em}
\end{table}

We address the horizon limitation by \emph{chaining}: decomposing a long-horizon task into a sequence of shorter sub-problems, each solvable within a single generation. After each generation, we take its final frame as the conditioning image for the next, extending the model's planning horizon across multiple generations. Together with EPBS, this forms \ours{} (\textbf{Ch}aining with \textbf{Ea}rly \textbf{P}lanning).

\para{Pivot selection.}
A valid pivot frame must satisfy two properties: (1)~the agent has made forward progress toward the goal, and (2)~the agent has not entered any constraint-violating cell.
We use the same trajectory extraction pipeline as our verifier to identify the agent’s final valid cell position. Among valid candidates, we select the one closest to the goal. If no candidate makes valid forward progress, chaining terminates.

\para{Compute budget.}
Each chain depth runs a full EPBS round, so total compute scales as $D \times B$ NFEs (max depth $D=3$).
In practice, most mazes require only $2$--$3$ chain steps, as each successful chain step covers $6-10$ maze cells.

\begin{figure}[!t]
    \centering
    \includegraphics[width=0.9\textwidth]{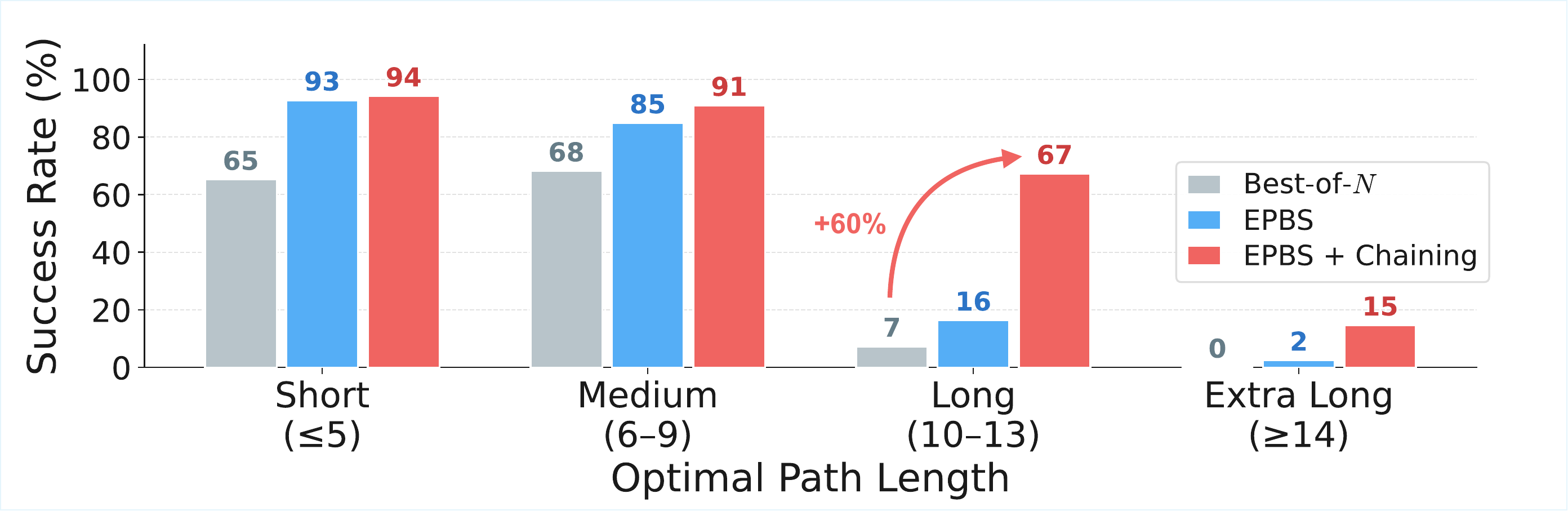}
    \caption{\textbf{Chaining extends reasoning beyond the single-generation horizon.} 
    Wan 2.2 success rates on Frozen Lake mazes by solution length.
    Chaining provides the most benefit on mazes where the solution length exceeds the generation window ($\geq 10$ steps).
    The largest gain is on long mazes, from $7\%$ to $67\%$ success rate with ChEaP.
    }
    \label{fig:chaining_results}
\end{figure}

\subsection{Results}
\label{subsec:chaining_results}

\Cref{tab:main_results} shows our full \ours{} results. 

For Wan 2.2 on size 6 mazes, \ours{} achieves 88.5\% pass@2, a 33.4$\%$ improvement over EPBS alone and more than double the best-of-$N$ rate of 43.6\%.
The gains are largest \textbf{precisely where EPBS is bottlenecked} by the generation horizon (\cref{fig:chaining_results}).
For Wan 2.2 on long mazes (path length 10--13), best-of-$N$ achieves 7.3\%, EPBS reaches 16.4\%, and \ours{} achieves 67.3\%---demonstrating that the model possesses local planning ability but cannot express full solutions in a single generation.
On extra-long mazes (14+ steps), chaining improves EPBS from 2.4\% to 14.6\%; the smaller gain reflects compounding errors across chains.

\section{What Breaks When Video Models Fail?}
\label{sec:failure_analysis}

Pass rate alone does not reveal \emph{why} video models fail on maze reasoning. 
To better understand the limits of video models, we categorize failures into three coarse groups: \textbf{constraint violations}, where the agent enters a forbidden region or otherwise breaks maze structure; \textbf{horizon-limited failures}, where the agent follows a plausible route prefix but fails to complete the task within the generation window; and \textbf{degenerate failures}, such as static agents, tracking failures, or severe output corruption. This decomposition lets us distinguish failures of \emph{structural adherence} from failures of \emph{sequential reach}. 

\subsection{Structural adherence degrades with difficulty}
\label{sec:failure-modes}

\begin{figure}[!t]
    \centering
    \includegraphics[width=\textwidth]{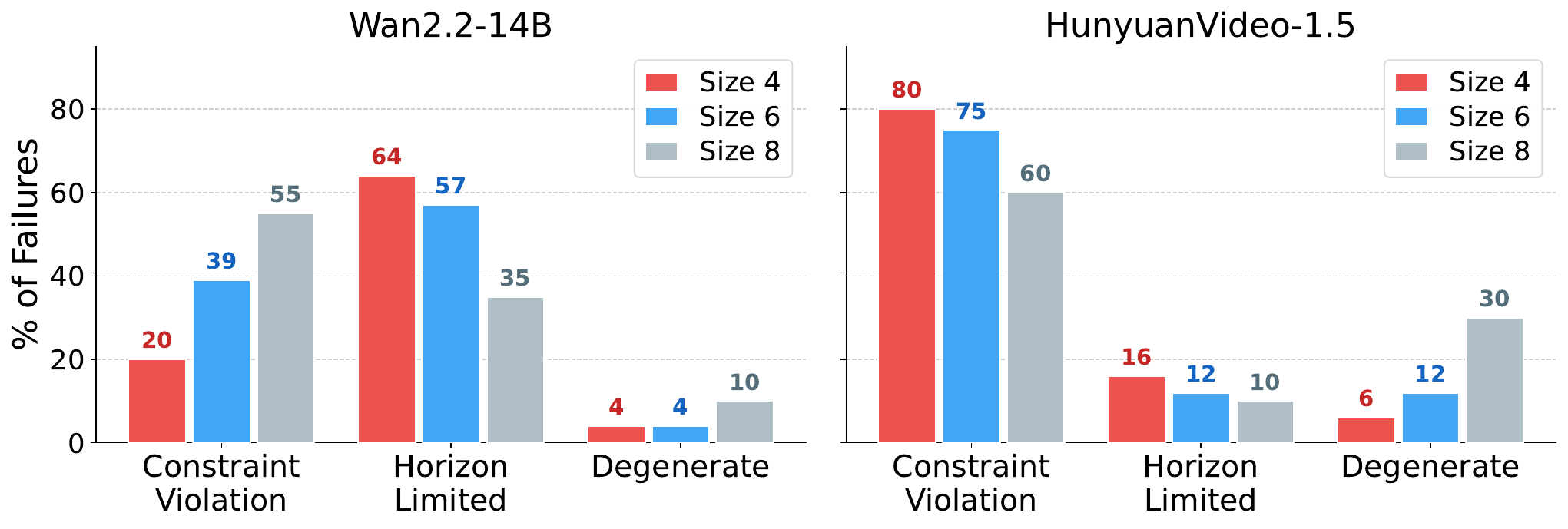}
    \caption{\textbf{Failure mode comparison.} Wan2.2 (left) goes from
    horizon-limited to constraint-dominated as maze size increases. HunyuanVideo (right)
    is constraint-dominated at \emph{all} sizes, suggesting that step distillation
    degrades structural adherence independently of horizon.
    }
    \label{fig:hunyuan-failures}
\end{figure}

We find that failure distributions across our two models differ significantly (\cref{fig:hunyuan-failures}).
Wan2.2 is dominated by horizon-limited failures on easy mazes (63.5\% on size~4), shifting to constraint-dominated only at size~10.
HunyuanVideo, by contrast, is constraint-dominated at \emph{all} sizes: 77.5\% of size-4 failures are constraint violations, compared to 32.5\% for Wan.
Even on small mazes where Wan almost never violates constraints, HunyuanVideo frequently moves the goal, enters lakes, or introduces illegal moves.
This pattern suggests that step distillation (8 steps vs.\ 40) degrades structural adherence independently of planning horizon.

\begin{figure}[!t]
    \centering
    \includegraphics[width=\textwidth]{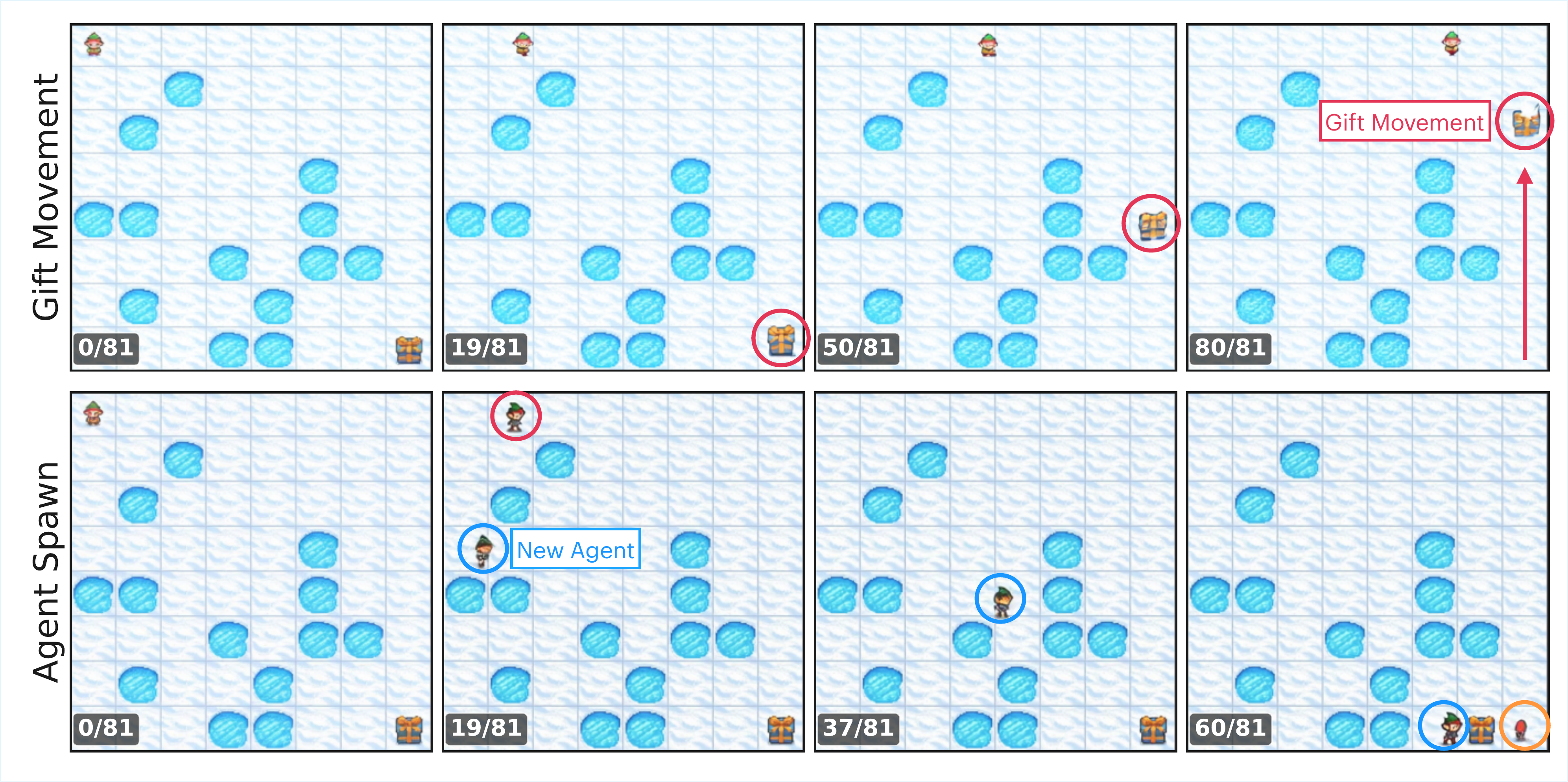}
    \caption{\textbf{The model ``cheats'' on hard mazes.} 
When the trajectory is too long to complete within the generation window, 
the model sacrifices structural adherence to fulfill the prompt. 
\emph{Top:} the gift teleports from the far corner to an adjacent cell, 
allowing the agent to ``solve'' the maze without traversing the full path. 
\emph{Bottom:} a second agent spawns near the goal and reaches it, 
while the original agent remains stranded.}
    \label{fig:cheating_examples}
\end{figure}

For Wan2.2, the shift from horizon-limited to constraint-dominated failures 
is consistent with the path length analysis in \Cref{sec:chaining}: 
as mazes require longer trajectories, the model faces a conflict between its 
early plan commitment and its generation horizon.
Rather than producing an incomplete but valid prefix, it often ``cheats'' to solve the maze by any means possible. 
The model moves the gift closer to the agent or spawns a second agent near the goal (\Cref{fig:cheating_examples}).
These behaviors are not random failures, but reflect a systematic breakdown under horizon pressure. 
The model preserves the \emph{intent} of the task (reaching the goal), 
but violates the underlying environment constraints to achieve it. In other words, 
when the required trajectory exceeds its effective generation horizon, the model 
prioritizes goal completion over structural fidelity.

We test this to the extreme with a set of controlled \emph{diagnostic} mazes, carefully designed to reveal patterns in the model's behaviors (\appref{sec:diagnostic-mazes}).
We find that the model struggles to adhere to constraints, especially on simple decoy mazes where the solution needs to go around a lake instead of directly to the goal.
This reflects a systematic bias for goal completion over constraint satisfaction.

\subsection{Implications for screening and chaining}
\label{subsec:failure_implications}

These failure modes clarify the roles of our two methods. \emph{Trajectory screening} is most useful when good trajectories exist in the candidate pool but are rare: it helps identify promising seeds early and avoid wasting compute on clearly invalid ones. \emph{Chaining}, by contrast, is most useful when failures are horizon-limited---that is, when the model can follow a valid route prefix but cannot complete the full trajectory within a single generation window. The detailed failure taxonomy supports this interpretation: smaller mazes contain many ``valid-prefix stall'' failures, which chaining can naturally address, whereas larger mazes show increasing constraint violations, which chaining alone cannot fix.

\section{Conclusion}
\label{sec:conclusion}

Our findings point to two complementary bottlenecks in video model reasoning.
Plans crystallize in the first few denoising steps, so inference-time scaling should prioritize exploring diverse candidates over refining individual ones; and strong local planning ability is bottlenecked by generation length, so extending the effective horizon---through longer native context windows, learned pivoting, or improved chaining---is equally critical for harder tasks.
Our work here focuses on mazes, but we believe the core principles are applicable more broadly.
Whether early commitment and horizon limitations manifest similarly in non-spatial reasoning modalities, and whether training can produce models that plan more reliably or over longer horizons, are important open questions.
More broadly, our results suggest that current video models are more capable reasoners than standard evaluations reveal; the bottleneck is less in what information models retain and more so in how we extract such knowledge.

\section*{Acknowledgements}
This work is supported by the National Science Foundation under Grant No. 2145198 and the Princeton First Year Fellowship to KN.
We also thank William Yang for helpful discussions and technical insights on the project, and Allison Chen and Esin Tureci for detailed feedback on the manuscript.

%% file: supp_content.tex
\section{Implementation Details}
\label{sec:impl-details}

\subsection{Trajectory Extraction Pipeline}
\label{sec:traj-extraction}

We extract cell-level trajectories from generated videos using a two-stage pipeline.

\para{Stage 1: Pixel-level tracking with SAM2.}
We use SAM2.1 (Hiera-Tiny variant) to track the elf sprite across video
frames~\cite{ravi_sam_2024}. The tracker is initialized in the first frame  with a bounding box derived from the known start-cell pixel region (from maze metadata). For each subsequent frame, SAM2 produces a segmentation mask from which
we extract the centroid $(c_x, c_y)$. We similarly track the goal (e.g.\ gift in Frozen Lake) to check for goal drift during generation.

\para{Stage 2: Centroid-to-cell mapping.}
Given the dimensions of the game board $(x_\text{min}, y_\text{min}, x_\text{max},
y_\text{max})$ and grid size $G$, we compute cell dimensions $w_\text{cell} = (x_\text{max}
- x_\text{min}) / G$ and $h_\text{cell} = (y_\text{max} - y_\text{min}) / G$. Each
centroid is mapped to a grid cell:
\begin{equation}
    \text{col} = \left\lfloor \frac{c_x - x_\text{min}}{w_\text{cell}} \right\rfloor,
    \quad
    \text{row} = \left\lfloor \frac{c_y - y_\text{min}}{h_\text{cell}} \right\rfloor
\end{equation}
with clamping to $[0, G-1]$. The trajectory is the ordered sequence of unique cells
visited.

\subsection{Success Criteria}
\label{sec:success-criteria}

A generated video is considered successful if the extracted trajectory satisfies:
(1) the trajectory ends at the goal cell (within grid tolerance);
(2) no intermediate cell falls on a hole or maze border (constraint violation). We use this permissive criterion to accept all valid solutions rather than comparing against ground-truth optimal paths. A common behavior on easier mazes is constraint violations \textbf{after} solving the maze: because these models are trained to produce smooth video throughout the 5-second window, they continue producing motion after the agent has already reached the goal. We handle this by truncating the trajectory at the first goal visit. Additionally, cases where the elf remains stationary or oscillates between two cells are classified as degenerate failures.

\subsection{Verifier Details}
\label{sec:verifier-details}

The verifier scores $\hat{x}_0$ predictions using background-difference motion
detection to estimate the agent’s trajectory, then combines goal progress with an obstacle penalty.

\para{Motion detection.}
In intermediate generations, obstacle cells (frozen lakes in Frozen Lake; traps or walls in VR-Bench) often flicker, which causes naive motion detection methods to fail.
For each frame, we compute the absolute pixel difference from the conditioning frame
(first frame), threshold at intensity 60, and find connected components with minimum area
50~pixels. The centroid of the largest component is taken as the agent position.

\para{Confidence scoring.}
Let $\mathcal{O}$ denote the set of obstacle cells (lakes, traps, or walls depending on the environment) and $F$ the number of video frames.
We localize the agent per-frame via the motion detection above, map each centroid to a maze cell, and derive two quantities: the final Manhattan distance to the goal $d(\mathrm{end}, \mathrm{goal})$ and the obstacle ratio $\lambda = |\{t_i : \mathrm{cell}(t_i) \in \mathcal{O}\}| / F$, i.e., the fraction of frames in which the agent’s centroid falls on an obstacle cell.
The confidence score is:
\begin{equation}
    c = 1 - \frac{d(\mathrm{end}, \mathrm{goal})}{d(\mathrm{start}, \mathrm{goal})} - \alpha \lambda,
\end{equation}
where $d(\cdot,\cdot)$ is Manhattan distance and $\alpha = 0.5$ controls the penalty for constraint violations.
Seeds are ranked by $c$, and only the top-$K$ are selected for full denoising.
This continuous score is robust to minor tracking noise while still penalizing clearly invalid trajectories.

\subsection{Generation Hyperparameters}
\label{sec:gen-hyperparams}

Table~\ref{tab:hyperparams} summarizes the generation settings. Both models use
image+text-to-video conditioning. We pad the condition images to 16:9 aspect ratio
($832{\times}480$ for Wan, 480p for Hunyuan) using centered black borders. Wan2.2-14B~\cite{wan_wan_2025}
uses the UniPC scheduler with shift 5.0. HunyuanVideo-1.5~\cite{wu_hunyuanvideo_2025} uses its native Euler
flow-matching scheduler with 8 step-distilled inference steps. We do not change any of the default hyperparameters for these models. 

\begin{table}[!t]
    \centering
    \caption{\textbf{Generation hyperparameters} for both video diffusion models.}
    \label{tab:hyperparams}
    \begin{tabular}{lcc}
        \toprule
        \textbf{Parameter} & \textbf{Wan2.2-14B} & \textbf{HunyuanVideo-1.5} \\
        \midrule
        Resolution            & $832 \times 480$              & 480p ($848 \times 480$)     \\
        Frame count           & 81                            & 121                         \\
        FPS                   & 16                            & 24                          \\
        Guidance scale        & 3.5                           & 6.0                         \\
        Scheduler             & UniPC                         & Euler (flow matching)       \\
        Denoising steps ($T$) & 40                            & 8 (step-distilled)          \\
        Compute & 4$\times$ Nvidia L40 & 4$\times$ Nvidia L40 \\
        \bottomrule
    \end{tabular}
\end{table}

\subsection{Text Prompts}
\label{sec:text-prompts}

We include the exact prompts we used to query the models for each benchmark.

\begin{tcolorbox}[colback=gray!5, colframe=gray!50, title=\textbf{Frozen Lake Prompt}]
\small
Animate the elf moving step by step toward the gift while carefully avoiding the icy frozen lake. Highlight the successful path and end with the elf touching the gift. There are no changes to the layout of the maze. No new lakes or characters appear. Static camera. No zoom. No pan. No glitches, noise, or artifacts.
\end{tcolorbox}
\begin{tcolorbox}[colback=gray!5, colframe=gray!50, title=\textbf{VR-Bench Trapfield Prompt Template}]
\small
Create a 2D animation based on the provided image of a maze. The \texttt{\{player\}} slides smoothly along the \texttt{\{floor\}} path, stopping perfectly on the \texttt{\{goal\}}. The \texttt{\{player\}} never slides into or crosses the \texttt{\{trap\}} (trap areas). The camera is a static, top-down view showing the entire maze.

\medskip
\textbf{Maze:} The maze paths are \texttt{\{floor\}}, and the trap areas are \texttt{\{trap\}}. The \texttt{\{player\}} moves to the goal position, represented by the \texttt{\{goal\}}. The \texttt{\{player\}} slides smoothly along the \texttt{\{floor\}} path. The \texttt{\{player\}} never slides into or crosses the \texttt{\{trap\}} of the maze. The \texttt{\{player\}} stops perfectly on the \texttt{\{goal\}}.

\medskip
\textbf{Scene:} No change in scene composition. No change in the layout of the maze. The \texttt{\{player\}} travels along the \texttt{\{floor\}} path without speeding up or slowing down.

\medskip
\textbf{Camera:} Static camera. No zoom. No pan. No glitches, noise, or artifacts.

\medskip
\textbf{Skins:}
\medskip

\begin{enumerate}[leftmargin=*, nosep, label=\textbf{\arabic*.}]
\item \texttt{player}=blue circle, \texttt{goal}=green circle, \texttt{trap}=red x, \texttt{floor}=white square
\item \texttt{player}=blue parka explorer \& penguin, \texttt{goal}=red flag, \texttt{trap}=blue water pool, \texttt{floor}=blue ice crystals
\item \texttt{player}=blue adventurer, \texttt{goal}=golden eagle emblem, \texttt{trap}=fiery explosion, \texttt{floor}=gray stone bricks
\item \texttt{player}=gray robot, \texttt{goal}=golden star, \texttt{trap}=blue crystal block, \texttt{floor}=silver metal plate
\end{enumerate}

\end{tcolorbox}

\begin{tcolorbox}[colback=gray!5, colframe=gray!50, title=\textbf{VR-Bench Maze Prompt Template}]
\small
Create a 2D animation based on the provided image of a maze. The \texttt{\{player\}} slides smoothly along the \texttt{\{floor\}} path, stopping perfectly on the \texttt{\{goal\}}. The \texttt{\{player\}} never slides or crosses into the \texttt{\{wall\}} areas of the maze. The camera is a static, top-down view showing the entire maze.

\medskip
\textbf{Maze:} The maze paths are \texttt{\{floor\}}, the walls are \texttt{\{wall\}}. The \texttt{\{player\}} moves to the goal position, represented by \texttt{\{goal\}}. The \texttt{\{player\}} slides smoothly along the \texttt{\{floor\}} path. The \texttt{\{player\}} never slides or crosses into the \texttt{\{wall\}} areas of the maze. The \texttt{\{player\}} stops perfectly on the \texttt{\{goal\}}.

\medskip
\textbf{Scene:} No change in scene composition. No change in the layout of the maze. The \texttt{\{player\}} travels along the \texttt{\{floor\}} path without speeding up or slowing down.

\medskip
\textbf{Camera:} Static camera. No zoom. No pan. No glitches, noise, or artifacts.

\medskip
\textbf{Skins:}
\medskip
\begin{enumerate}[leftmargin=*, nosep, label=\textbf{\arabic*.}]
\item \texttt{player}=red circle, \texttt{goal}=green square, \texttt{wall}=light blue square, \texttt{floor}=white square
\item \texttt{player}=white rabbit, \texttt{goal}=orange carrots, \texttt{wall}=gray rock, \texttt{floor}=green grass tiles
\item \texttt{player}=blue robot head, \texttt{goal}=green and yellow tile, \texttt{wall}=black brick wall, \texttt{floor}=gray decorative tile
\item \texttt{player}=anime schoolgirl, \texttt{goal}=green square, \texttt{wall}=gray stone wall, \texttt{floor}=wooden floor tiles
\item \texttt{player}=green circle, \texttt{goal}=red circle, \texttt{wall}=blue potion bottle, \texttt{floor}=white square
\end{enumerate}

\end{tcolorbox}

\section{EPBS Sensitivity Analysis}
\label{sec:sensitivity}

We perform more in depth analysis of EPBS on Wan2.2-14B, sweeping our hyperparameters and justifying our method even when taking into account extra NFEs from VAE decode time.

\subsection{Ablation on Probe Step $\tau$}
\label{sec:ablation-tau}

\begin{figure}[t]
    \centering
    \includegraphics[height=0.36\textheight]{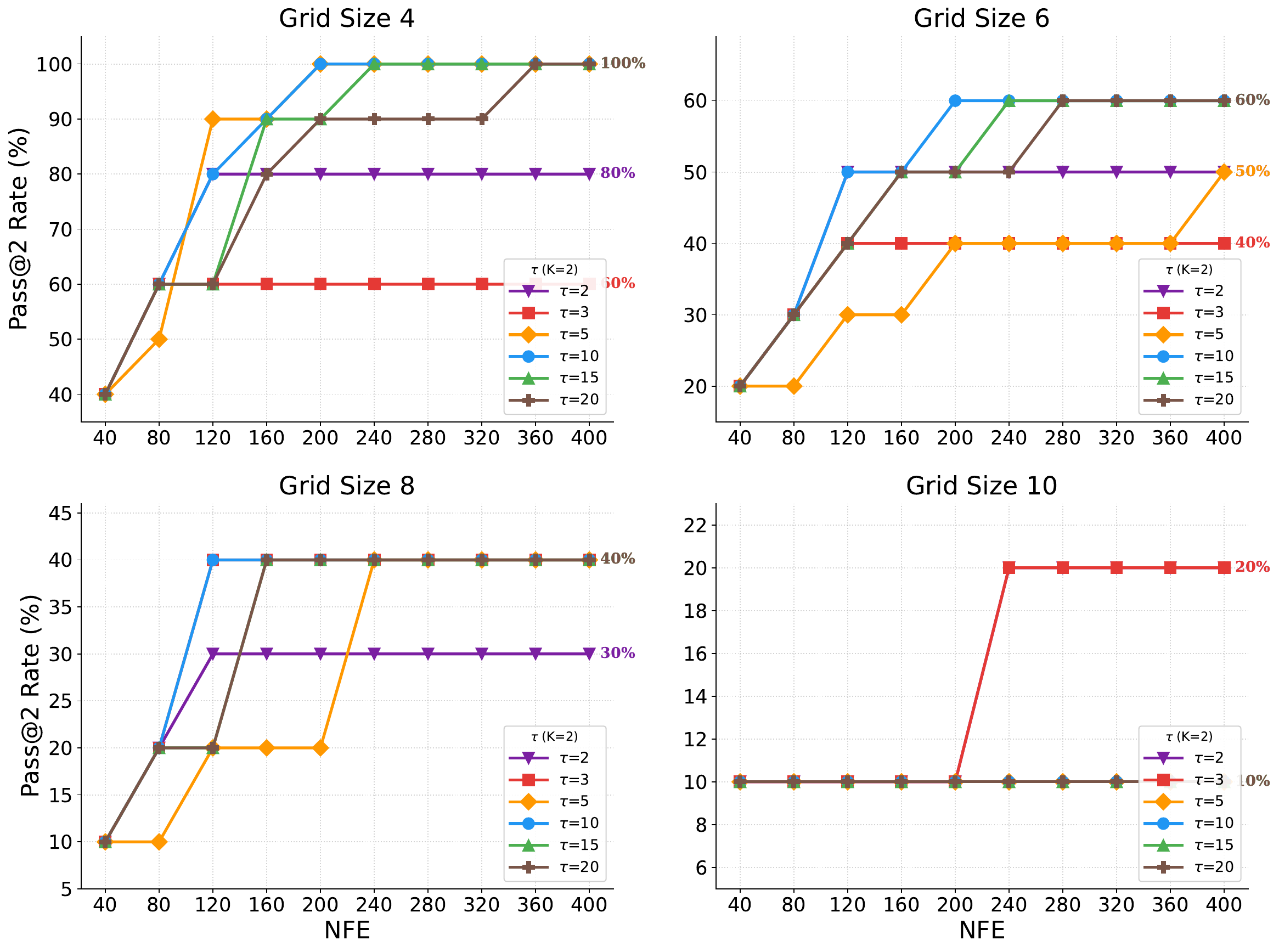}
    \caption{\textbf{Probe step sensitivity.} Pass@2 vs.\ NFE budget for probe steps
    $\tau \in \{2, 3, 5, 10, 15, 20\}$ at beam size $K=2$, across four grid sizes.
    All configurations use shared seed pools for fair comparison. $\tau = 5$ provides
    the best efficiency trade-off for size 4; larger mazes are less sensitive to $\tau$.}
    \label{fig:sweep-tau}
    \vspace{-2em}
\end{figure}

Figure~\ref{fig:sweep-tau} shows EPBS accuracy as a function of NFE budget for six probe
step values $\tau \in \{2, 3, 5, 10, 15, 20\}$ with beam size $K=2$ and a fixed random seed generator. 
We use a randomly selected subset of 10 mazes per size. 
For size~4, $\tau=5$ reaches peak accuracy at the lowest NFE; $\tau=2$ underperforms because the $\hat{x}_0$ prediction at that stage carries insufficient
trajectory information. For sizes 6--10, sensitivity to $\tau$ decreases, though $\tau=10$ tends to be a safe choice and very
early probing ($\tau=2$) consistently underperforms. The optimal $\tau$ reflects a
trade-off: too early, and the prediction lacks discriminative signal; too late, and the
probing cost approaches full generation.

\subsection{Ablation on Beam Size $K$}
\label{sec:ablation-K}

\begin{figure}[!b]
    \centering
    \includegraphics[height=0.36\textheight]{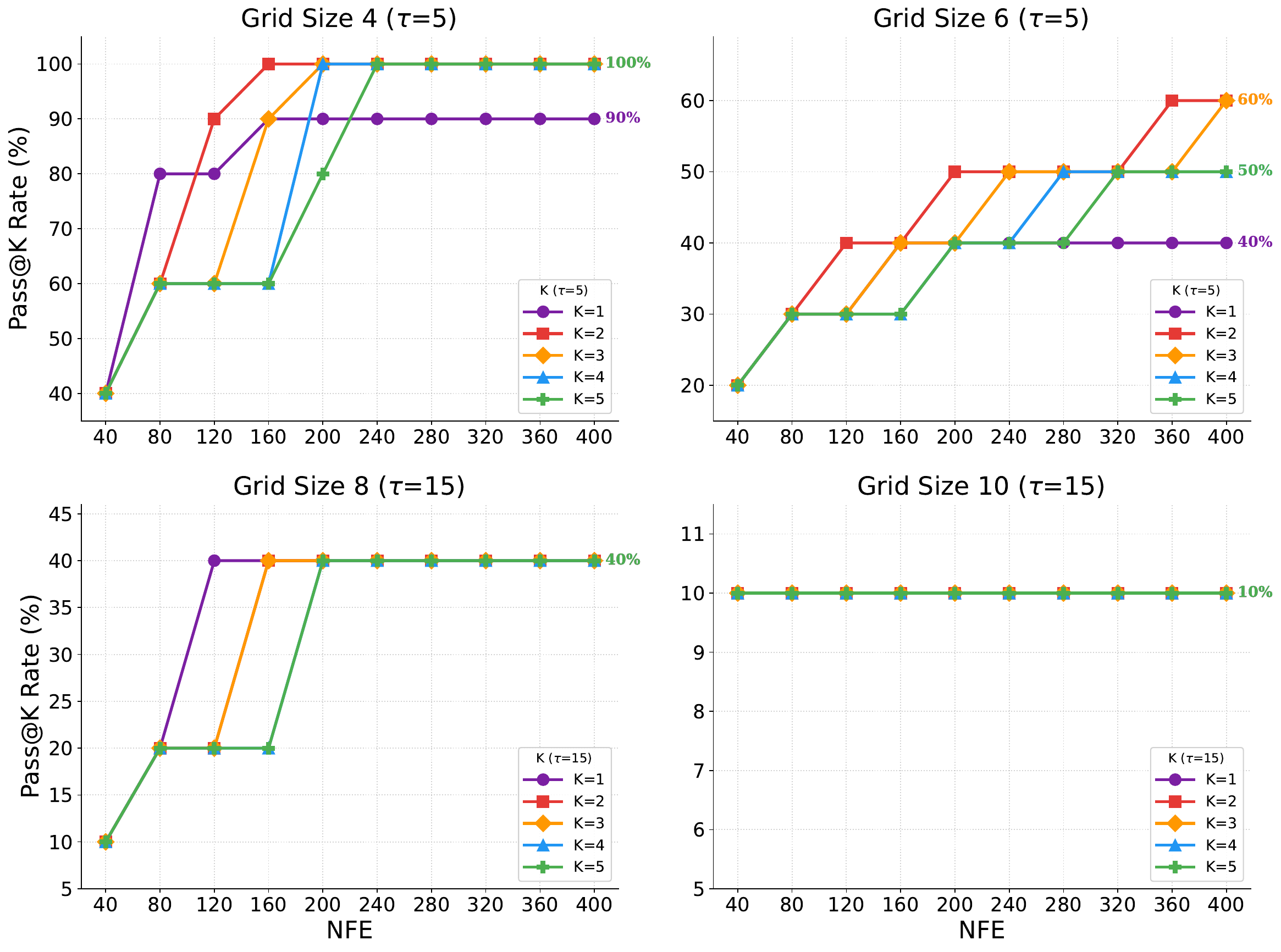}
    \caption{\textbf{Beam size ablation.} Pass@$K$ vs. NFE budget for beam sizes
    $K \in \{1, 2, 3, 4, 5\}$ with fixed probe step ($\tau=5$ for sizes 4--6,
    $\tau=15$ for sizes 8--10). $K{=}2$ provides the best trade-off: it reaches
    peak accuracy at lower budgets than $K{=}1$ without the probe-count penalty of
    larger $K$ values.}
    \label{fig:beam-ablation}
\end{figure}

The beam size $K$ controls how many top-scoring seeds are fully denoised after probing.
Figure~\ref{fig:beam-ablation} shows the effect of varying $K$ from 1 to 5 while holding
$\tau$ fixed. At low budgets, higher $K$ values are constrained: since each full completion costs $T{-}\tau$ steps, a limited budget leaves no room for the extra completions that larger $K$ demands, so all configurations reduce to completing the same small number of seeds and their curves overlap. As budget
increases, $K{=}2$ consistently reaches peak accuracy earliest: on size~4, $K{=}2$
achieves 100\% by NFE~160, while $K{=}1$ plateaus at 90\%. On size~6, $K{=}2$ reaches
60\% vs.\ 40\% for $K{=}1$. Larger beam sizes ($K{\geq}3$) provide no additional benefit
and can even reduce accuracy at moderate budgets by consuming NFE on completions rather
than probing more seeds. This confirms that $K{=}2$ optimally balances exploration
(probing diverse seeds) against exploitation (completing promising candidates).

\subsection{Wall-Clock Comparison}
\label{sec:wallclock}

\begin{table}[t]
    \centering
    \caption{\textbf{Wall-clock comparison at matched accuracy.} For each grid size, we
    report EPBS accuracy and wall-clock time at NFE~=~120, alongside the baseline NFE and
    time required to reach the same accuracy. Speedup = baseline time / EPBS time.
    Sizes 4--6 use $\tau{=}5$; sizes 8--10 use $\tau{=}15$.
    All wall-clock estimates include every pipeline component (see text for breakdown).}
    \label{tab:wallclock}
    \begin{tabular}{lccccccc}
        \toprule
        & & \multicolumn{2}{c}{\textbf{EPBS @ NFE~120}} &
          \multicolumn{2}{c}{\textbf{Baseline to match}} & \\
        \cmidrule(lr){3-4}\cmidrule(lr){5-6}
        \textbf{Size} & $\tau$ & Acc (\%) & Time (min) & NFE needed & Time (min) & Speedup \\
        \midrule
        4  &  5 & 88.2 & 25.9 & ${>}400^\dagger$ & 82.0 & 3.2$\times$ \\
        6  &  5 & 42.3 & 25.9 & 320              & 65.6 & 2.5$\times$ \\
        8  & 15 & 16.9 & 22.6 & 160              & 32.8 & 1.5$\times$ \\
        10 & 15 &  8.5 & 22.6 & 120              & 24.6 & 1.1$\times$ \\
        \bottomrule
    \end{tabular}
    \vspace{2pt}

    {\small $^\dagger$Baseline reaches 89.5\% at NFE~400 (closest match to EPBS's 88.2\%).}
\end{table}

We compare wall-clock time at matched accuracy levels rather than matched NFE budgets,
since EPBS trades additional NFE for better seed selection. All timings are measured on
4$\times$L40 GPUs with FSDP and sequence parallelism.

Each EPBS probe consists of $\tau$ denoising steps, VAE decoding of the $\hat{x}_0$
prediction, and background-difference verifier scoring. Each completion runs the remaining
$T{-}\tau$ denoising steps and VAE decodes the final video. SAM2 trajectory evaluation
(0.1~min per seed) is applied to every completed seed in both EPBS and baseline.
A full baseline generation (40 steps plus VAE decode) takes 8.1~min. The denoising cost
per step is approximately 11.2s, with a fixed overhead of ${\sim}0.3$~min for VAE
decoding and verifier scoring per probe, and ${\sim}0.3$~min for VAE decoding per
completion.

For sizes 4--6 ($\tau{=}5$), each probe costs 1.2~min and each completion costs 6.9~min.
At NFE~=~120, EPBS performs 10~probes ($10{\times}1.2{=}12.0$~min), 2~completions
($2{\times}6.9{=}13.7$~min), and 2~SAM2 evaluations ($2{\times}0.1{=}0.2$~min), totaling
25.9~min. For sizes 8--10 ($\tau{=}15$), each probe costs 3.1~min and each completion
costs 5.0~min. At NFE~=~120, EPBS performs 4~probes ($4{\times}3.1{=}12.4$~min),
2~completions ($2{\times}5.0{=}10.0$~min), and 2~SAM2 evaluations ($0.2$~min), totaling
22.6~min---fewer probes are needed because each probe is more expensive but also more
informative.

Table~\ref{tab:wallclock} shows that at NFE~=~120, EPBS achieves accuracy that the baseline
requires NFE~=~120--400 to match, yielding \textbf{1.1--3.2$\times$ wall-clock speedup.} 
The speed advantage is largest on small mazes where EPBS's screening is most effective.

\FloatBarrier
\section{Extended Analysis}
\label{sec:extended}

\subsection{Cross-Model Early Plan Commitment}
\label{sec:crossmodel-lockin}

\begin{figure}[t]
    \centering
    \includegraphics[width=0.68\textwidth]{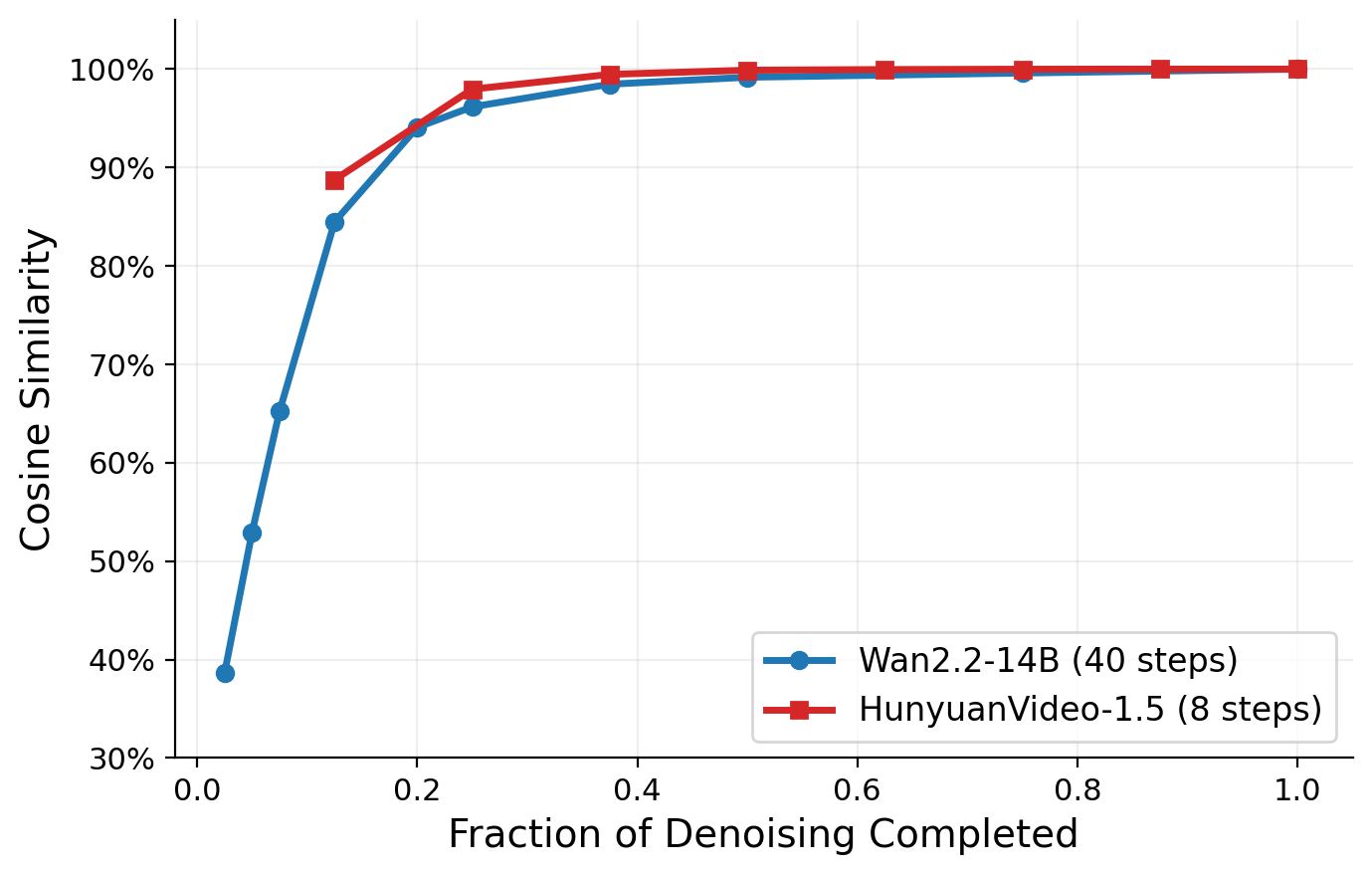}
    \caption{\textbf{Cross-model early plan commitment across all maze sizes.}
    Mean trajectory convergence $\mathcal{C}$ between intermediate $\hat{x}_0$ predictions and the final video, plotted against normalized schedule fraction.
    We compute $\mathcal{C}$ using the cosine similarity between motion-energy maps, exactly as in the main text.
    Both Wan2.2-14B ($T{=}40$) and HunyuanVideo-1.5 ($T{=}8$) commit to a trajectory within the first 10--15\% of their denoising schedules, after which convergence largely plateaus.
    Despite using different schedulers and step counts, the normalized convergence profiles are similar, suggesting that early plan commitment is a structural property of video diffusion rather than a model-specific artifact.}
    \label{fig:crossmodel}
\end{figure}

The main paper establishes early plan commitment on $4{\times}4$ Frozen Lake mazes using Wan2.2-14B. Here we show that the same phenomenon holds across all maze sizes and across both models.

Following the main text, we measure trajectory convergence using the cosine similarity between intermediate and final \emph{motion-energy maps}. For each decoded $\hat{x}_0^{(t)}$, we accumulate background-difference motion over frames into a grid-aligned energy map $\mathbf{M}^{(t)}$, flatten it to $\mathbf{m}^{(t)}$, and compare it to the final prediction $\mathbf{m}^{(T)}$:
\[
\mathcal{C}(\text{step } t)
= \frac{\mathbf{m}^{(t)} \cdot \mathbf{m}^{(T)}}
       {\|\mathbf{m}^{(t)}\| \, \|\mathbf{m}^{(T)}\|}.
\]
As in the main text, this metric is more robust than discrete cell overlap on larger grids, where small tracking noise can introduce low-level activity in irrelevant cells.

Figure~\ref{fig:crossmodel} plots mean trajectory convergence against normalized schedule fraction. Wan2.2-14B and HunyuanVideo-1.5 follow the same qualitative pattern: convergence rises sharply in the earliest portion of the denoising schedule and then plateaus.
HunyuanVideo starts at a higher absolute value because its step-distilled schedule compresses much more progress into each step, but after normalization the two profiles align closely.
This shows that early plan commitment is not specific to one architecture, scheduler, or step count; it appears to be a general property of video diffusion denoising.
\subsection{HunyuanVideo-1.5 Analysis}
\label{sec:hunyuan-deep-dive}

\begin{figure}[!t]
    \centering
    \includegraphics[width=0.72\textwidth]{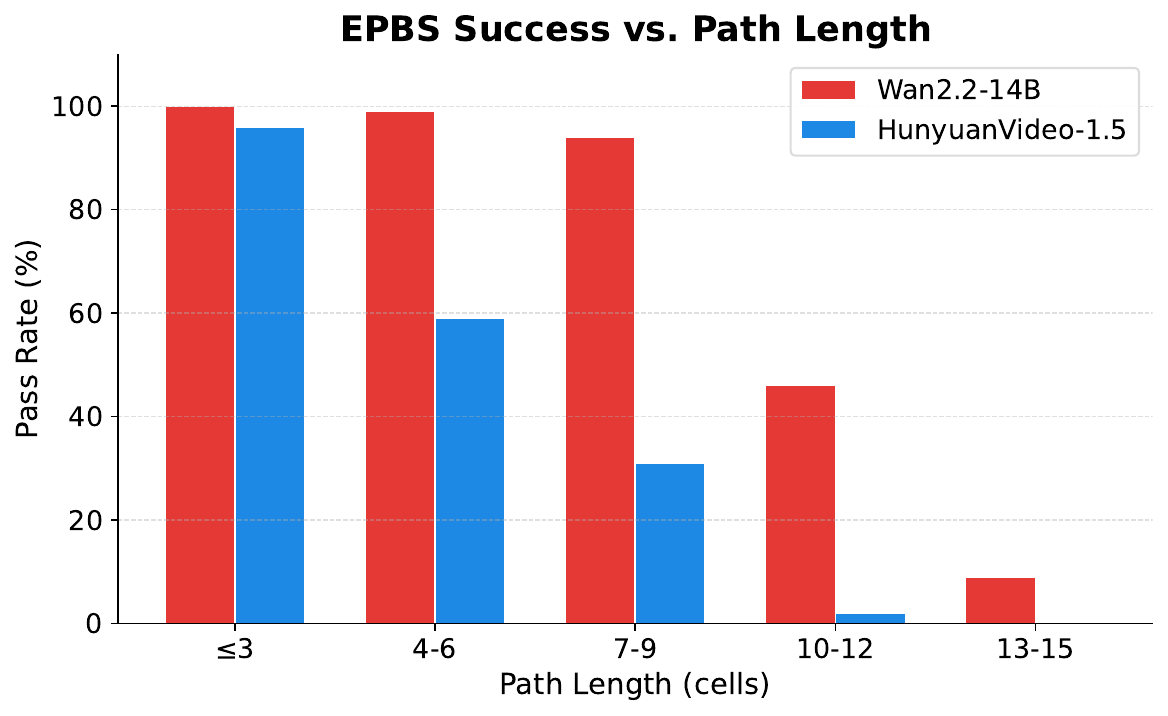}
    \caption{\textbf{Path length is the dominant difficulty axis for both models.}
    HunyuanVideo's effective planning horizon is shorter: both models solve short paths
    reliably but diverge sharply beyond 7 cells.}
    \label{fig:hunyuan-path-length}
\end{figure}

In this section, we apply analysis to HunyuanVideo-1.5 of similar depth that we applied to Wan2.2 in the main paper. We use the HunyuanVideo-1.5 Step-Distilled ($T=8$) model, using probe step $\tau=2$ for size~4 and $\tau=3$ for sizes~6 and~8, with beam size $K=2$. 
\para{Path length dominates difficulty.}
As with Wan, path length is the primary difficulty axis, with Pearson correlations of
$r{=}{-}0.40$ (size~4), $r{=}{-}0.77$ (size~6), and $r{=}{-}0.77$ (size~8). Figure~\ref{fig:hunyuan-path-length} compares the two models on path length performance. HunyuanVideo hits the horizon wall
earlier: Wan maintains 46\% at path length 10--12 and 9\% at 13--15, while HunyuanVideo
drops to near zero beyond 9 cells, confirming that size-10 evaluation would yield near-zero results.

\para{Verifier reliability.}
Despite HunyuanVideo's weaker generation quality, the $\hat{x}_0$ verifier remains
informative. Top-2 precision is 46.4\% on size~4 (vs.\ 26.0\% random baseline, a 1.8$\times$ gain ---i.e., the verifier's top-2 selections contain successful seeds 1.8$\times$ more often than random),
18.1\% on size~6 (1.7$\times$), and 10.8\% on size~8 (1.3$\times$). The reduced gain 
relative to Wan (2.2--5.5$\times$) from \cref{tab:verifier_precision} is consistent with HunyuanVideo's noisier 8-step
$\hat{x}_0$ predictions carrying less discriminative trajectory information per step.

\FloatBarrier
\subsection{Diagnostic Maze Variants}
\label{sec:diagnostic-mazes}

\begin{figure}[!htb]
    \centering
    \includegraphics[width=\textwidth]{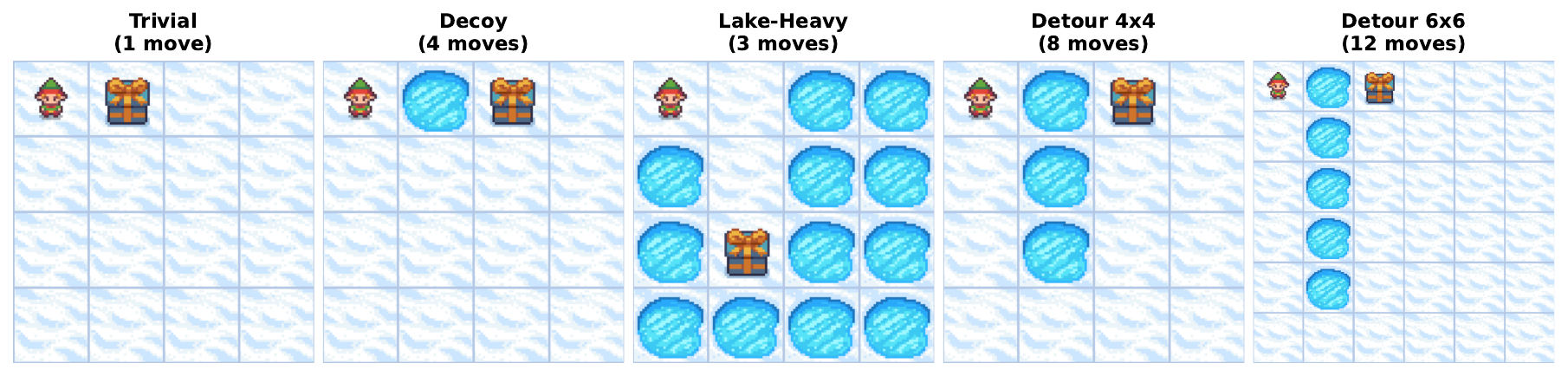}
    \caption{\textbf{Diagnostic maze variants.} From left to right: \emph{Trivial} (1--2
    moves, ceiling test), \emph{Decoy} (goal visually adjacent but blocked by lake),
    \emph{Lake-Heavy} ({>}75\% lake, single narrow corridor), and
    \emph{Detour} (Manhattan distance~2 but 8--12 move path around a lake wall).}
    \label{fig:diagnostic-mazes}
\end{figure}

To stress-test the verifier and probe the model's failure modes under controlled conditions,
we design four categories of diagnostic mazes (Figure~\ref{fig:diagnostic-mazes}) that each
isolate a specific challenge. We evaluate Wan2.2-14B with EPBS ($\tau{=}5$, $K{=}2$, budget~400). Table~\ref{tab:diagnostic} reports two metrics: the
\emph{per-seed} success rate (what fraction of individual generations solve the maze) and the
\emph{EPBS} success rate (whether EPBS finds at least one correct solution among all seeds for each maze). Detour and decoy mazes are particularly informative for validating the verifier design.
In both categories, the goal is Manhattan distance ${\leq}2$ from the start, so seeds
that take an illegal shortcut through the lake \emph{score highly on the progress
component} of our verifier---they end very close to the goal. The fact that EPBS
nonetheless rejects these seeds and selects valid detours demonstrates that the
constraint penalty $\alpha \lambda$ in Eq.~2 is essential: without it, the verifier
would systematically prefer the illegal shortcuts.

\para{Trivial mazes} (1--2 moves) serve as a ceiling test. Even on the easiest possible
mazes, 40\% of seeds fail---producing gift movement or wrong-path errors---confirming that
the model's generation process is inherently stochastic and that seed selection adds value
even when the planning problem itself is trivial. EPBS solves all~4 trivial mazes.

\para{Decoy mazes} place the goal visually adjacent to the start (one cell away) but block
the direct path with a lake tile, requiring a 4--5 step detour. This is the hardest
category despite the short path length: only 6\% of seeds succeed, and EPBS solves just
1 of~4 mazes. The dominant failure is lake entry (55\%)---the model overwhelmingly takes
the one-step shortcut through the forbidden cell rather than navigating around it. Unlike
detour mazes, where the difficulty stems from path length exceeding the planning horizon,
decoy mazes fail for a purely \emph{perceptual} reason: the model cannot resist the
visual shortcut even when the valid path is well within its planning capacity.
\begin{figure}[!htb]
    \centering
    \includegraphics[width=0.52\textwidth]{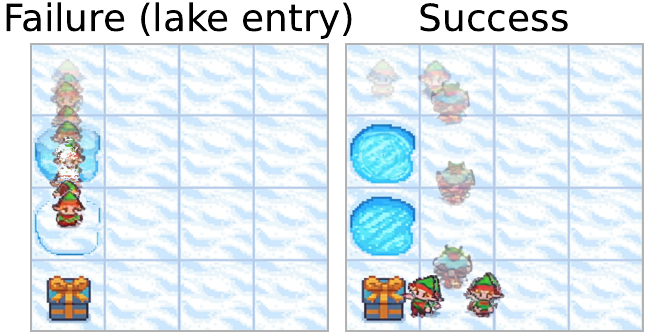}
    \caption{\textbf{Decoy maze: failure vs.\ success.} Ghost-trail visualizations on a
    4$\times$4 decoy maze where the goal is visually adjacent but blocked.
    \emph{Left}: the model beelines toward the visible goal through the lake
    (lake entry). \emph{Right}: the rare seed that navigates the 5-step
    detour around the obstruction.}
    \label{fig:decoy-comparison}
\end{figure}

\para{Lake-Heavy mazes} ({>}75\% lake tiles) force navigation through a single narrow
corridor. Despite the extreme constraint density, EPBS solves all~4 mazes (69\% per-seed).
The few failures split between lake entry and gift movement, indicating
that dense obstacles do not fundamentally break the model---consistent with the main paper's finding that obstacle density has near-zero correlation with difficulty.

\begin{figure}[!htb]
    \centering
    \includegraphics[width=0.52\textwidth]{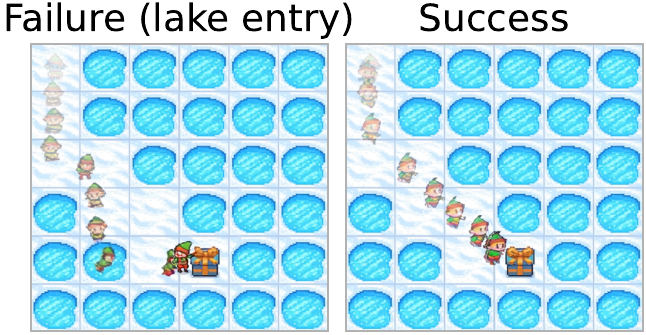}
    \caption{\textbf{Lake-heavy maze: failure vs.\ success.} Ghost-trail visualizations on a
    6$\times$6 lake-heavy maze ({>}75\% lake). \emph{Left}: the agent follows the correct
    corridor but cuts through a single lake cell at step~5 (lake entry).
    \emph{Right}: the EPBS-selected seed navigates the full 7-step corridor without
    constraint violations.}
    \label{fig:lakeheavy-comparison}
\end{figure}

\para{Detour mazes} are the most revealing category. The start and goal are only Manhattan
distance~2 apart, but a lake wall forces the agent to take a long path of 8~moves (size~4)
or 12~moves (size~6) around the obstruction. When faced with this conflict between visual
proximity and actual path length, the model's dominant failure mode is striking: rather
than planning the long detour, it hallucinates the goal sliding closer to the agent (gift
movement in 80\% of size-4 failures). The model appears to ``choose'' to modify the
environment rather than solve the harder planning problem. On size~4, only 29\% of seeds
navigate the detour correctly, yet EPBS solves both mazes---the verifier's constraint
penalty rejects the illegal shortcuts and surfaces the rare seeds that respect the maze
topology. On size~6 (12-move detour), no seeds succeed at all; the path exceeds the
model's effective planning horizon, and EPBS cannot select what the model never generates.

\begin{figure}[!htb]
    \centering
    \includegraphics[width=0.52\textwidth]{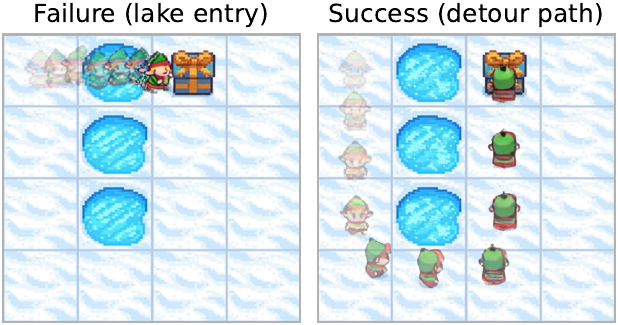}
    \caption{\textbf{Detour maze: failure vs.\ success.} Ghost-trail visualizations on the
    same 4$\times$4 detour maze. \emph{Left}: a typical failure---the agent walks straight
    toward the goal through the lake (constraint violation). \emph{Right}: the EPBS-selected
    success---the agent navigates the 8-step detour around the lake wall. The verifier's
    constraint penalty correctly rejects the shortcut and surfaces the rare seed that
    respects the maze topology.}
    \label{fig:detour-comparison}
\end{figure}

Figure~\ref{fig:detour-comparison} illustrates this contrast directly: on the same maze,
most seeds take the visually obvious straight-line path through the lake, while the
EPBS-selected seed navigates the full 8-step detour.

\begin{table}[!b]
    \centering
    \caption{\textbf{Diagnostic maze results.} \emph{Per-seed} reports the fraction of
    individual generations that solve the maze; \emph{EPBS} reports the fraction of mazes
    for which EPBS finds at least one correct solution across all seeds. Failure modes are
    assigned by priority (gift movement~{$>$}~lake entry~{$>$}~wrong path).}
    \label{tab:diagnostic}
    \begin{tabular}{llcccl}
        \toprule
        \textbf{Category} & \textbf{Size} & \textbf{Path}
        & \textbf{Per-Seed} & \textbf{EPBS}
        & \textbf{Dominant Failure Mode} \\
        \midrule
        Trivial   & 4 & 1--2  & 60\% & 100\% & gift mvmt.\ (33\%), wrong path (33\%) \\
        Decoy       & 4--6 & 4--5  & 6\%  & 25\%  & lake entry (55\%), wrong path (17\%) \\
        Lake-Heavy  & 4--6 & 3--7  & 69\% & 100\% & lake entry (50\%), gift mvmt.\ (25\%) \\
        Detour    & 4 & 8     & 29\%  & 100\% & gift movement (80\%) \\
        Detour    & 6 & 12    & 0\%  & 0\%   & gift mvmt.\ (36\%), lake entry (27\%) \\
        \bottomrule
    \end{tabular}
\end{table}

Together, these diagnostics reveal two distinct failure regimes. \emph{Detour} failures
are horizon-limited: the model cannot sustain a plan over 12+ steps, even when it
``knows'' the correct direction. \emph{Decoy} failures are perception-limited: the model
has sufficient planning capacity but is overwhelmed by the visual salience of the nearby
goal. EPBS is effective against both regimes when correct seeds exist in the pool, but
it cannot compensate when the model systematically fails to generate any valid trajectory.

\subsection{Trajectory diversity across seeds}
\label{subsec:supp_trajectory_diversity}

\begin{figure}[h]
    \centering
    \includegraphics[width=\linewidth]{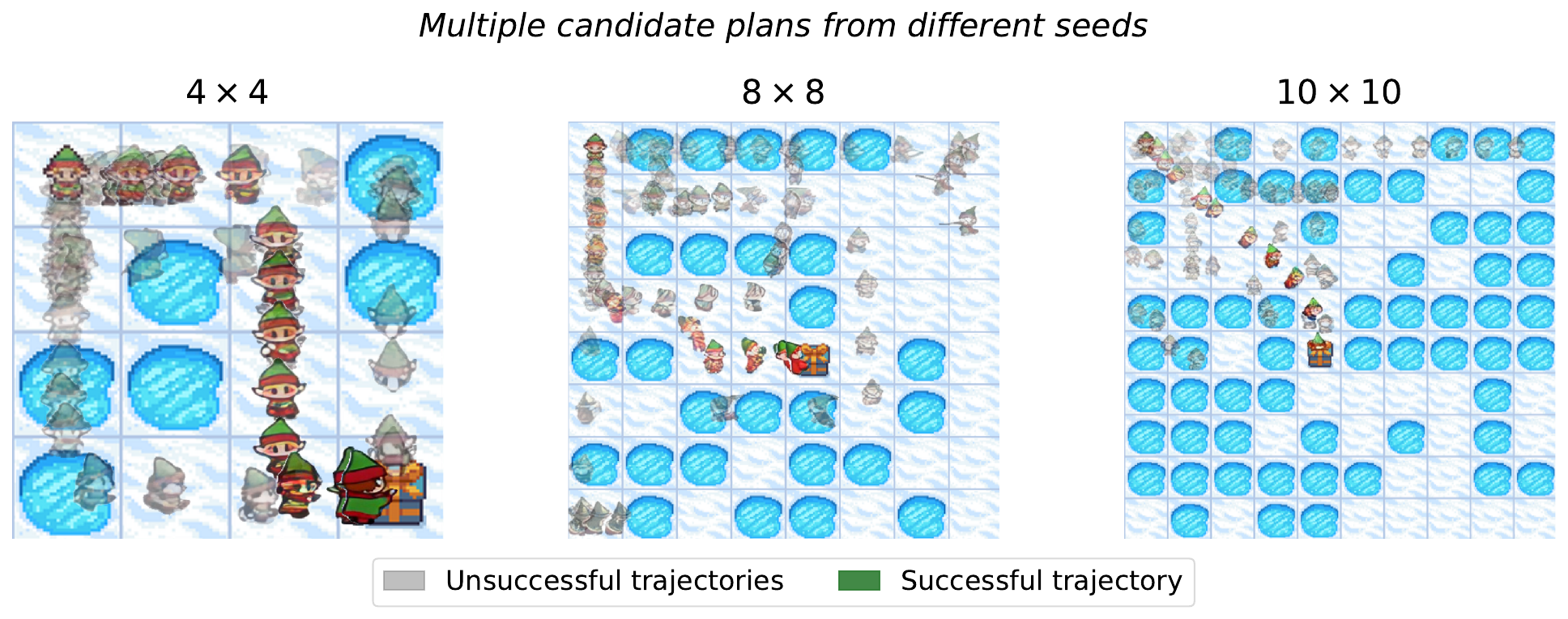}
    \caption{\textbf{Diversity of candidate plans under different noise seeds.}
    We overlay trajectories extracted from multiple samples of the same maze.
    Incorrect trajectories are shown in gray with reduced opacity, while a successful trajectory is shown in color.
    Despite sharing identical conditioning, different seeds produce diverse motion plans.
    Importantly, these plans are distinguishable early in the denoising process, suggesting that inference-time compute should be allocated toward selecting promising trajectories rather than refining all candidates.}
    \label{fig:trajectory_diversity}
\end{figure}

While \Cref{fig:plan_lockin} shows that individual trajectories stabilize early, it does not capture the diversity of plans explored across different noise seeds.
In \Cref{fig:trajectory_diversity}, we visualize multiple sampled trajectories overlaid on the same maze.
We observe substantial diversity in candidate plans, with many trajectories failing due to suboptimal routing or constraint violations.
Crucially, these trajectories are already distinguishable at early denoising steps, indicating that the model explores a space of candidate solutions before committing to a final plan.
This observation motivates allocating inference-time compute toward selecting among early plans rather than refining individual trajectories.

\FloatBarrier
\section{Additional Qualitative Examples}
\label{sec:qualitative}

\setlength{\floatsep}{4pt}
\setlength{\textfloatsep}{6pt}

We provide qualitative examples of early commitment, successful chaining, and failure
modes. In all galleries, the conditioning frame (maze image) is shown on the left,
followed by decoded $\hat{x}_0$ predictions at denoising steps $t \in \{1, 3,
5, 15, 40\}$, with the final generated video on the right (average frame visualized).

\subsection{Early Commitment Gallery}
\label{sec:gallery-commitment}

Each row below shows a ghost-trail visualization of a single seed's decoded
$\hat{x}_0$ prediction at denoising steps $\tau \in \{2, 5, 15, 20\}$ and the
final generated video. The trajectory is visible by $\tau{=}5$ and remains
stable through later steps, confirming early plan commitment.
We distinguish \emph{norm} mazes (goal at the far corner, maximizing path length) from \emph{vary} mazes (randomly placed goal, often admitting shorter solutions). HunyuanVideo-1.5 uses a step-distilled schedule with $T{=}8$ total steps, so its rows show $\hat{x}_0$ at $\tau \in \{1, 3, 5, 7\}$ instead.

{%
\setlength{\intextsep}{2pt}%
\setlength{\abovecaptionskip}{2pt}%
\setlength{\belowcaptionskip}{0pt}%

\begin{figure}[H]\centering
\includegraphics[width=0.88\textwidth]{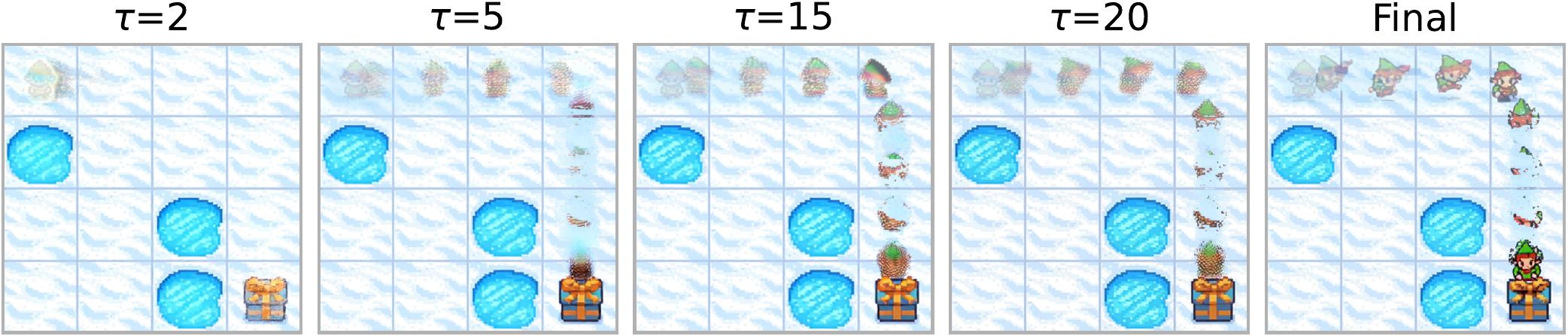}
\caption{\textbf{Wan2.2, size~4 (norm).} The full solution path is already visible
in the $\tau{=}5$ prediction; subsequent steps only sharpen the rendering without altering
the planned route.}\end{figure}

\begin{figure}[H]\centering
\includegraphics[width=0.88\textwidth]{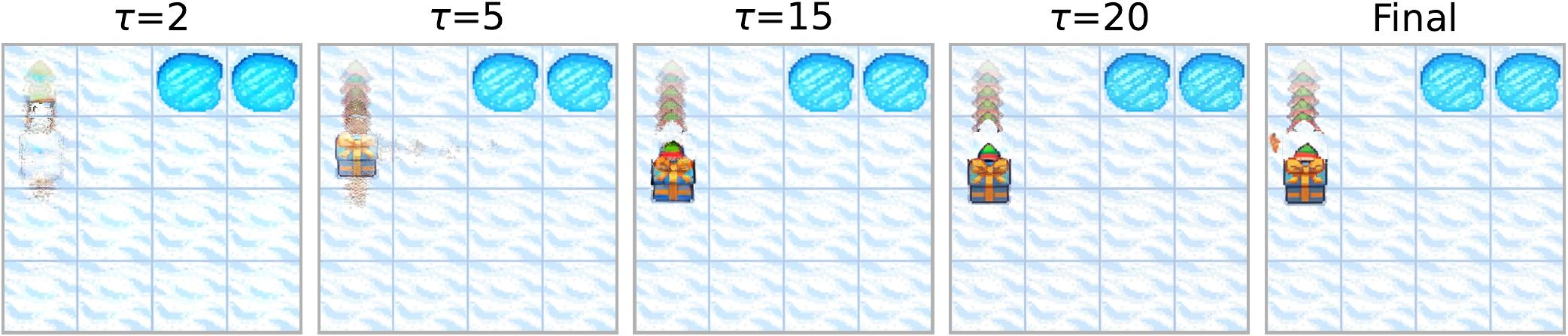}
\caption{\textbf{Wan2.2, size~4 (vary).} The planned route is committed to by $\tau{=}5$ and maintains it through the final
video.}\end{figure}

\begin{figure}[H]\centering
\includegraphics[width=0.88\textwidth]{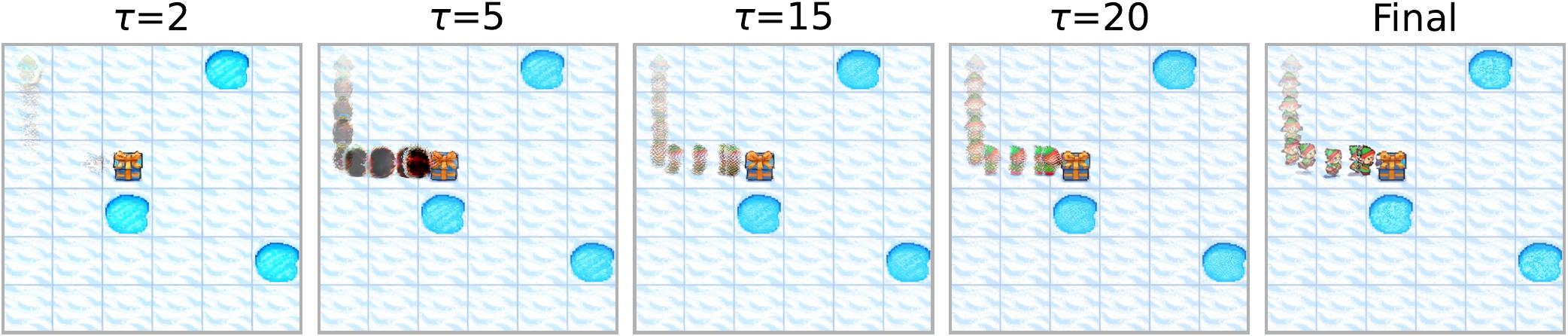}
\caption{\textbf{Wan2.2, size~6 (vary).} A longer maze requiring more steps.
Despite the increased path complexity, the overall route direction is locked in by
$\tau{=}5$.}\end{figure}

\begin{figure}[H]\centering
\includegraphics[width=0.88\textwidth]{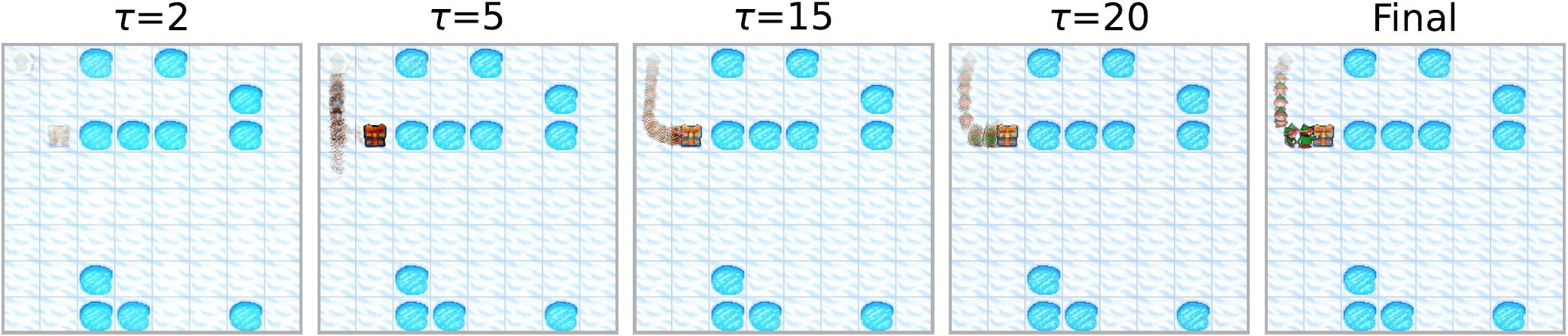}
\caption{\textbf{Wan2.2, size~8 (vary).} Size~8 maze with varied lake ratio.
The early prediction captures the general trajectory shape, though fine-grained cell-level
details continue to refine through later steps.}\end{figure}

\begin{figure}[H]\centering
\includegraphics[width=0.88\textwidth]{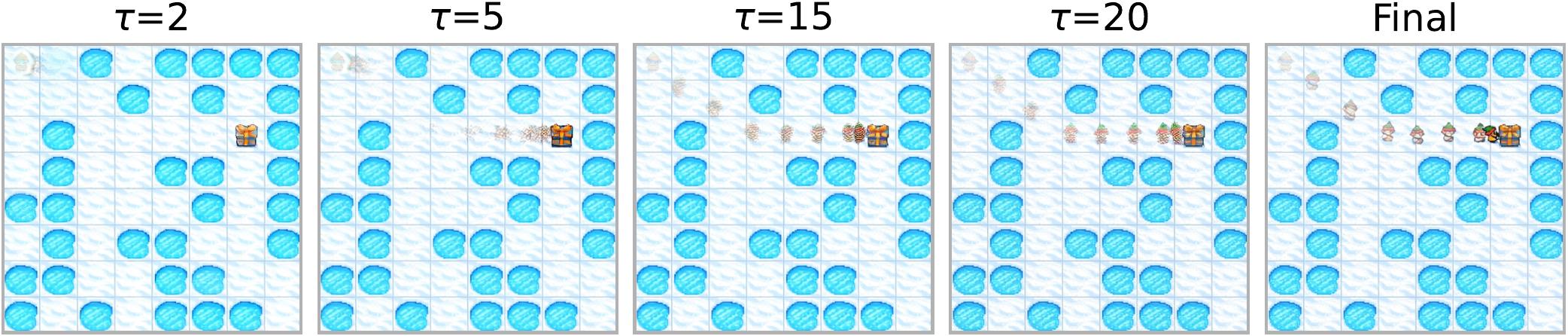}
\caption{\textbf{Wan2.2, size~8 (vary), second example.} The ghost
trail at $\tau{=}5$ already traces a path consistent with the final video, demonstrating
that early commitment holds across different maze instances.}\end{figure}

\begin{figure}[H]\centering
\includegraphics[width=0.88\textwidth]{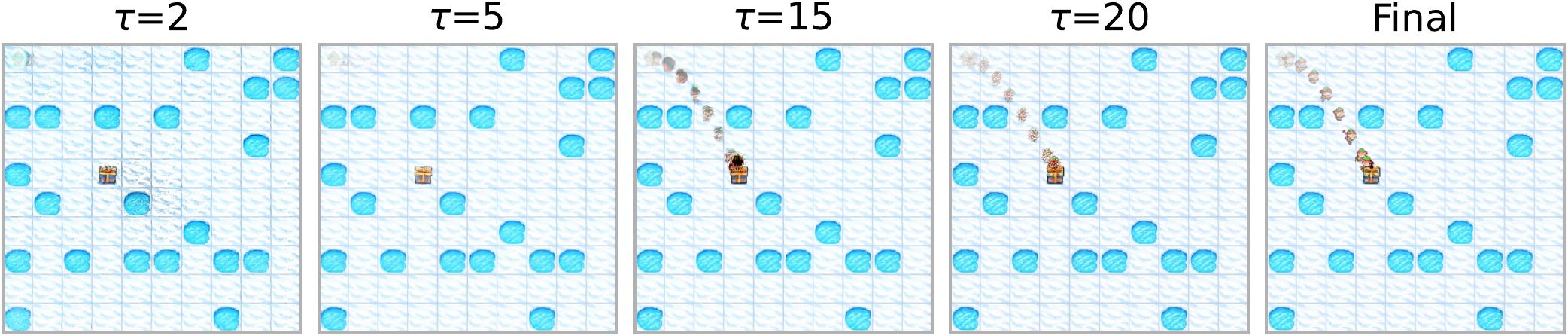}
\caption{\textbf{Wan2.2, size~10 (vary).} The largest maze size. Even for this
10$\times$10 grid, the model's planned trajectory is apparent by step~5 of denoising.}\end{figure}

\begin{figure}[H]\centering
\includegraphics[width=0.88\textwidth]{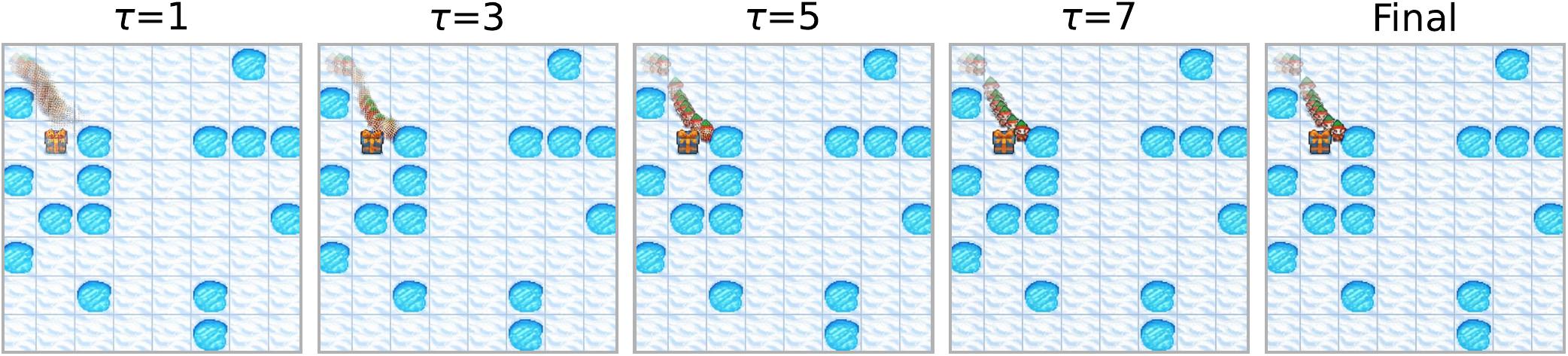}
\caption{\textbf{HunyuanVideo-1.5, size~8 (vary).} HunyuanVideo uses a
step-distilled schedule ($T{=}8$). Despite the shorter schedule, the trajectory is
committed by step~3 and remains stable, mirroring the early commitment seen in
Wan2.2.}\end{figure}

\begin{figure}[H]\centering
\includegraphics[width=0.88\textwidth]{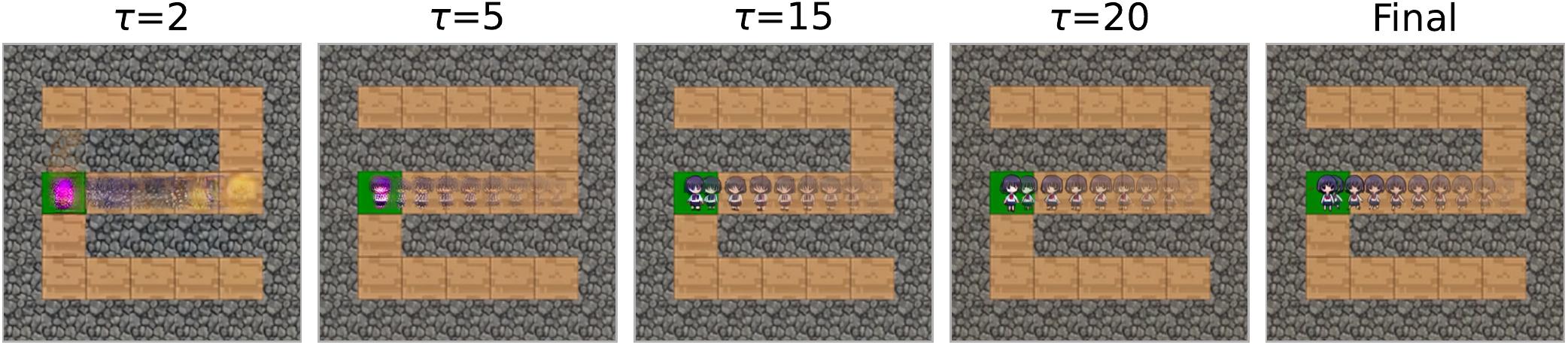}
\caption{\textbf{VR-Bench, maze\_4 (easy).} Early plan commitment on a VR-Bench maze with
a distinct brown/tan texture. The trajectory direction is already visible at $\tau{=}5$
and refines through later steps, showing that commitment generalizes beyond the Frozen Lake
visual style to procedurally generated maze textures.}\end{figure}

\begin{figure}[H]\centering
\includegraphics[width=0.88\textwidth]{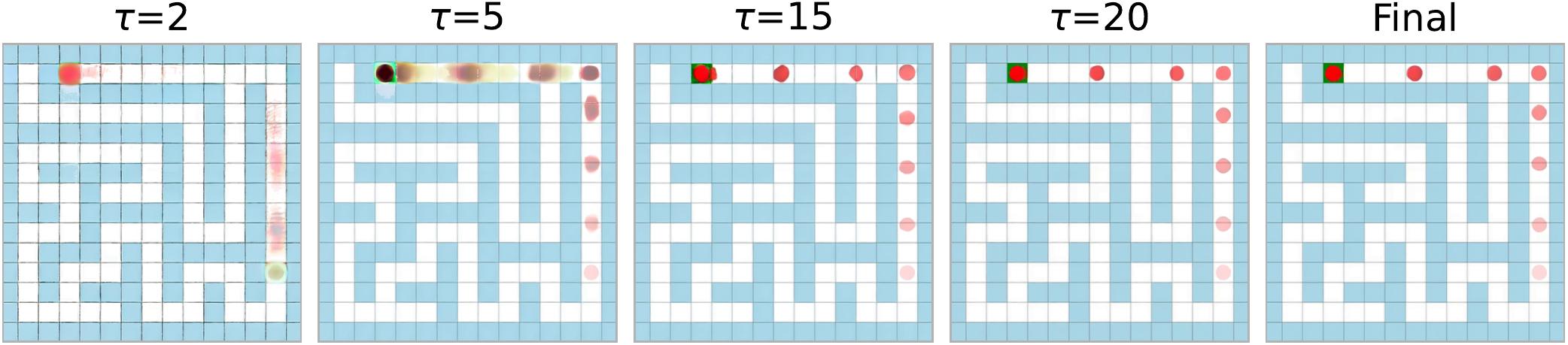}
\caption{\textbf{VR-Bench, maze\_1 (hard).} A harder VR-Bench puzzle on the blue/teal
maze\_1 texture. Despite higher path complexity, the $\hat{x}_0$ prediction at $\tau{=}5$
already captures the coarse trajectory shape, which persists through to the final
generation.}\end{figure}

\begin{figure}[H]\centering
\includegraphics[width=0.88\textwidth]{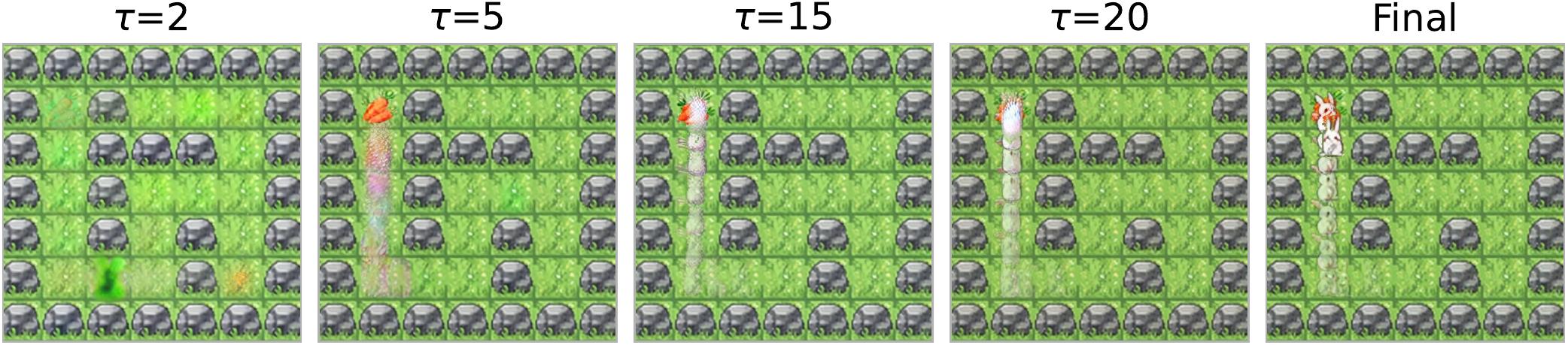}
\caption{\textbf{VR-Bench, maze\_2 (easy).} The purple/blue maze\_2 texture. The ghost
trail at $\tau{=}2$ is largely noise, but by $\tau{=}5$ the planned path is clearly
committed and matches the final output.}\end{figure}

}

\subsection{ChEaP Gallery}
\label{sec:gallery-chaining}

Each row shows the ghost-trail of each chain segment, with the first panel as the pivot seed followed by the chain segment and the final stitched video.

{%
\setlength{\intextsep}{2pt}%
\setlength{\abovecaptionskip}{2pt}%
\setlength{\belowcaptionskip}{0pt}%

\begin{figure}[H]\centering
\includegraphics[width=0.88\textwidth]{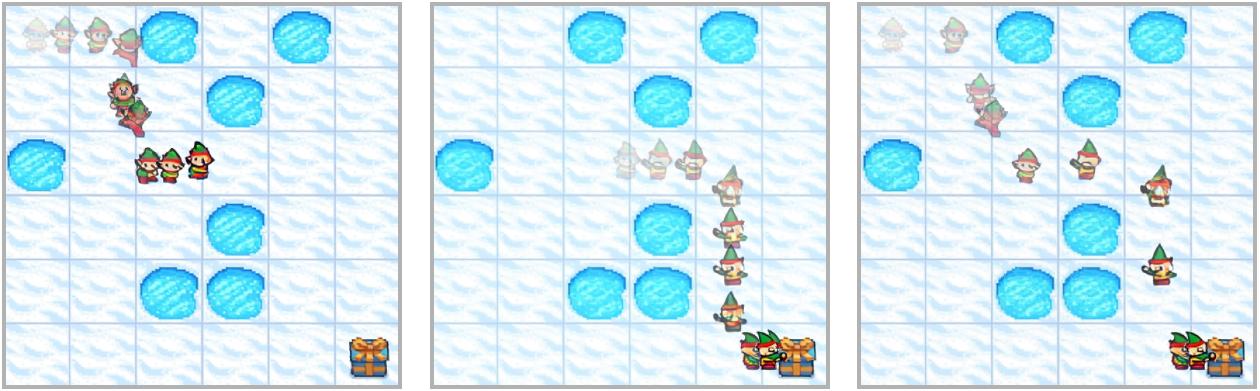}
\caption{\textbf{Wan2.2 chain, size~6 (norm), lake~20, maze~002.} Two-segment chain.
}\end{figure}

\begin{figure}[H]\centering
\includegraphics[width=0.88\textwidth]{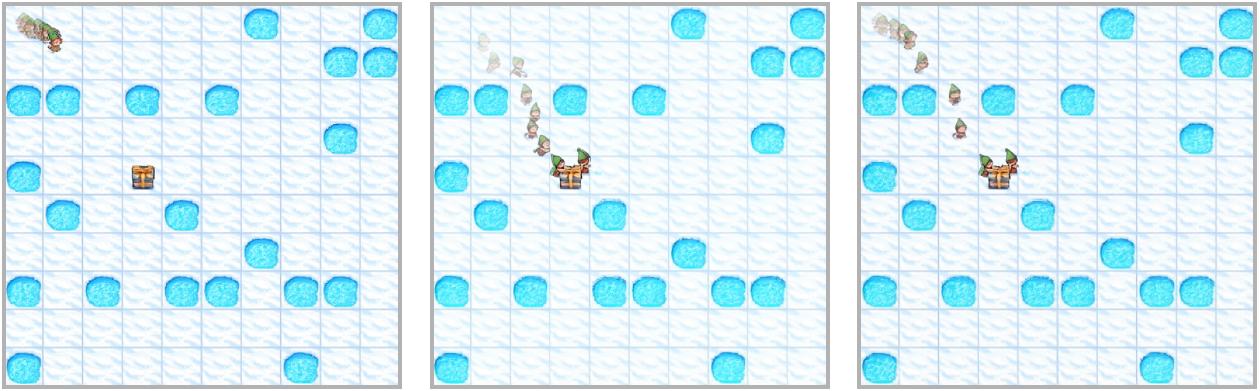}
\caption{\textbf{Wan2.2 chain, size~10 (vary), lake~20, maze~002.} Size~10 maze with
varied lake ratio. The longer solution path necessitates chaining.}\end{figure}

\begin{figure}[H]\centering
\includegraphics[width=0.88\textwidth]{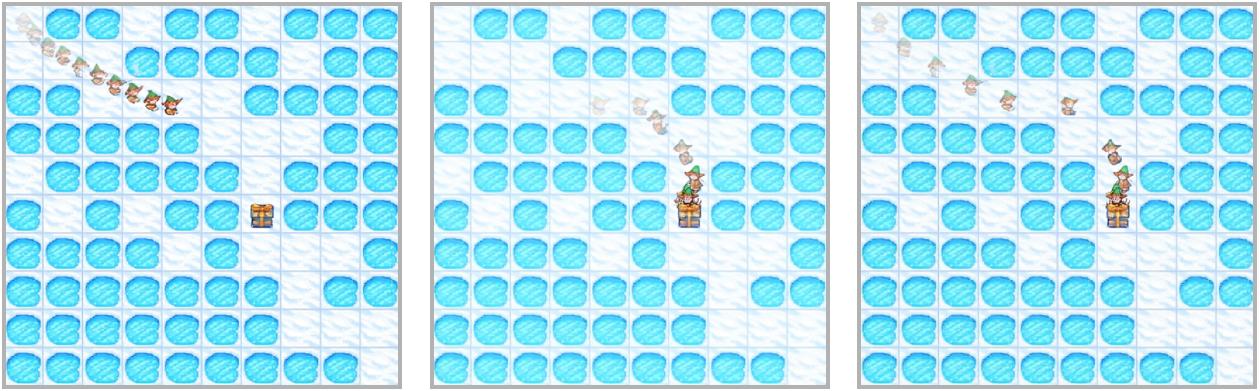}
\caption{\textbf{Wan2.2 chain, size~10 (vary), lake~80, maze~003.} High lake density leaves fewer safe cells, creating a narrow corridor that chaining navigates successfully.}\end{figure}

\begin{figure}[H]\centering
\includegraphics[width=0.88\textwidth]{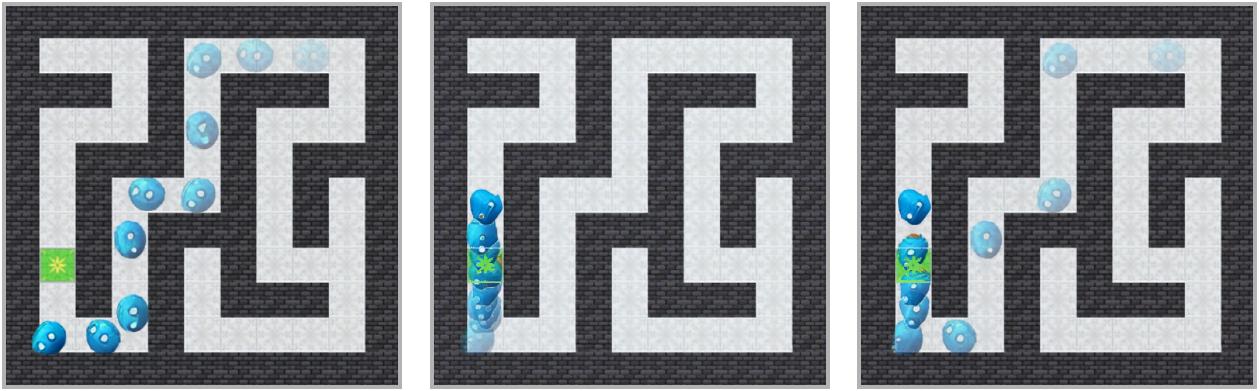}
\caption{\textbf{Wan2.2 chain, VR-Bench maze\_3\_medium, puzzle~0004.} Chaining applied
to a VR-Bench maze with a different visual texture than Frozen Lake. The agent navigates
a medium-difficulty procedurally generated maze across multiple segments.}\end{figure}

\begin{figure}[H]\centering
\includegraphics[width=0.88\textwidth]{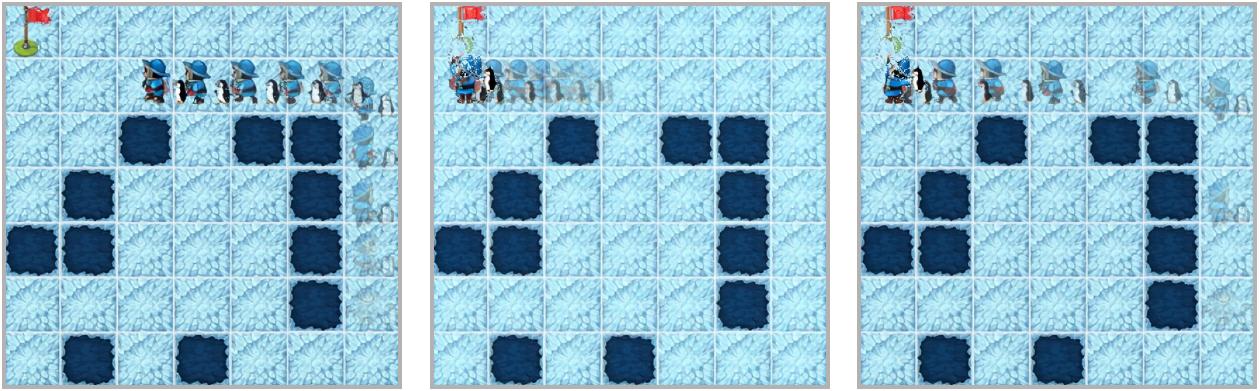}
\caption{\textbf{Wan2.2 chain, VR-Bench trapfield\_2\_medium, puzzle~0002.} A trapfield
environment where the agent must avoid trap cells (analogous to lakes). Chaining enables
completion of longer trapfield paths.}\end{figure}

\begin{figure}[H]\centering
\includegraphics[width=0.88\textwidth]{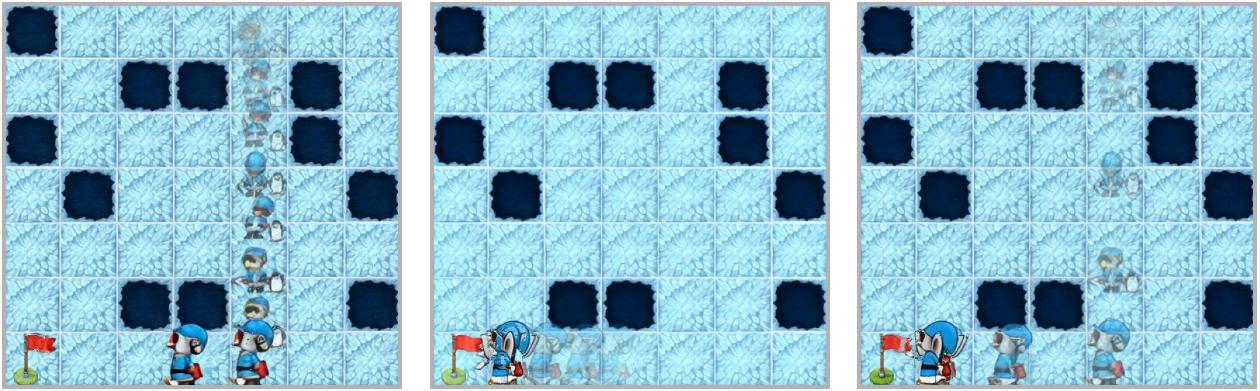}
\caption{\textbf{Wan2.2 chain, VR-Bench trapfield\_2\_medium, puzzle~0005.} Another
trapfield example showing successful multi-segment navigation through a different
puzzle layout.}\end{figure}

\begin{figure}[H]\centering
\includegraphics[width=0.88\textwidth]{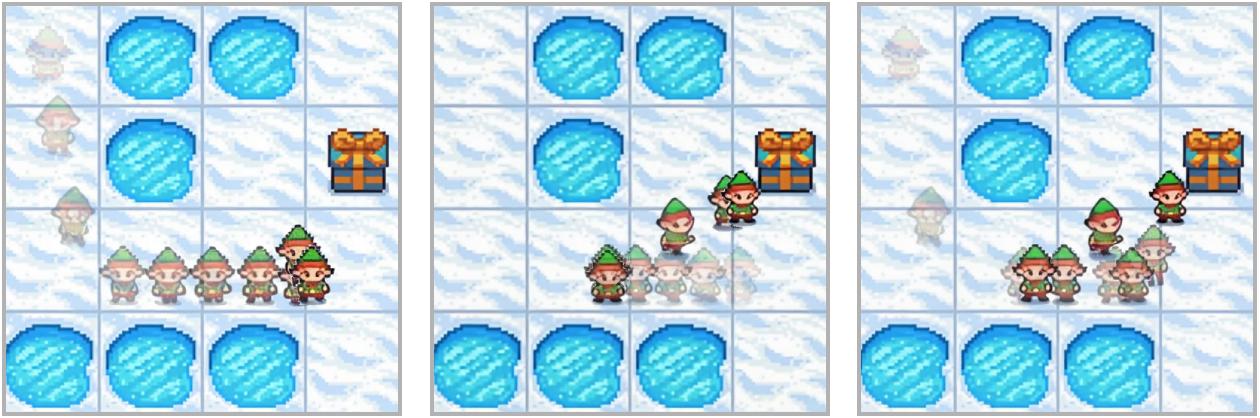}
\caption{\textbf{HunyuanVideo-1.5 chain, size~4 (vary), lake~65, maze~004.} Chaining
with HunyuanVideo's step-distilled schedule ($T{=}8$). The pivot and chain segments
are stitched to form a complete solution.}\end{figure}

\begin{figure}[H]\centering
\includegraphics[width=0.88\textwidth]{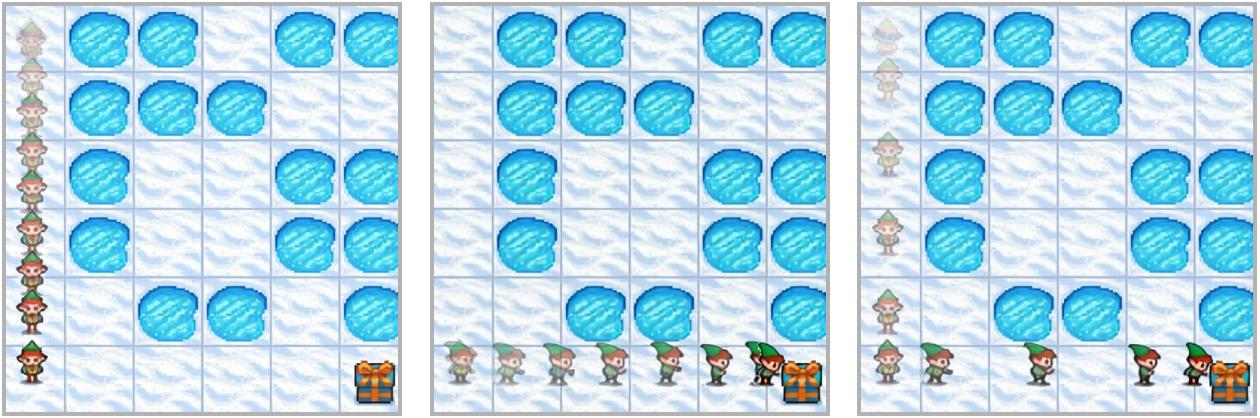}
\caption{\textbf{HunyuanVideo-1.5 chain, size~6 (norm), lake~65, maze~004.} A size~6
maze with 65\% lake density. Chaining extends HunyuanVideo's effective planning horizon
beyond a single generation.}\end{figure}

\begin{figure}[H]\centering
\includegraphics[width=0.88\textwidth]{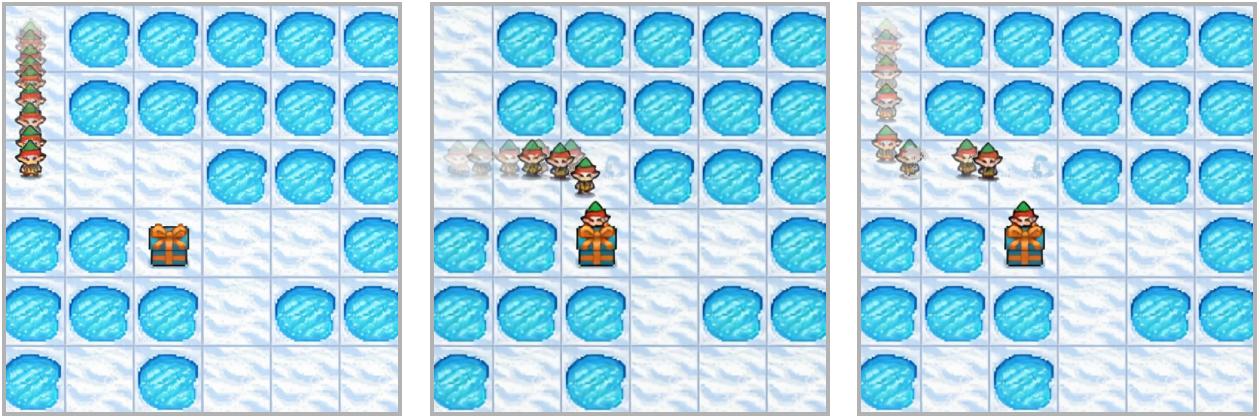}
\caption{\textbf{HunyuanVideo-1.5 chain, size~6 (vary), lake~80, maze~009.}
Despite the constrained environment, chaining produces a valid multi-segment solution.}\end{figure}
}

\subsection{Failure Mode Gallery}
\label{sec:gallery-failures}

Ghost-trail visualizations of representative failures from both models and domains. 
Left two panels: Wan 2.2; right two panels: HunyuanVideo-1.5.

{%
\setlength{\intextsep}{2pt}%
\setlength{\abovecaptionskip}{2pt}%
\setlength{\belowcaptionskip}{0pt}%

\begin{figure}[H]\centering
\includegraphics[width=0.88\textwidth]{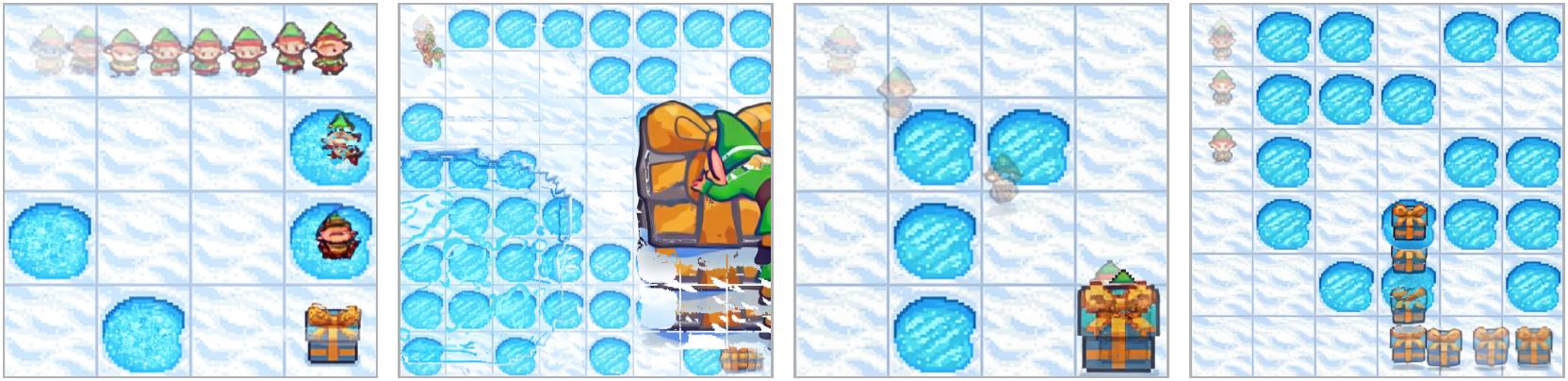}
\caption{\textbf{Constraint violations --- Frozen Lake.} Ghost-trail visualizations showing
two sub-types: \emph{lake entry} (the elf crosses into a frozen-lake cell, violating the
environment constraint) and \emph{gift movement} (the goal tile shifts position during
generation).}\end{figure}

\begin{figure}[H]\centering
\includegraphics[width=0.88\textwidth]{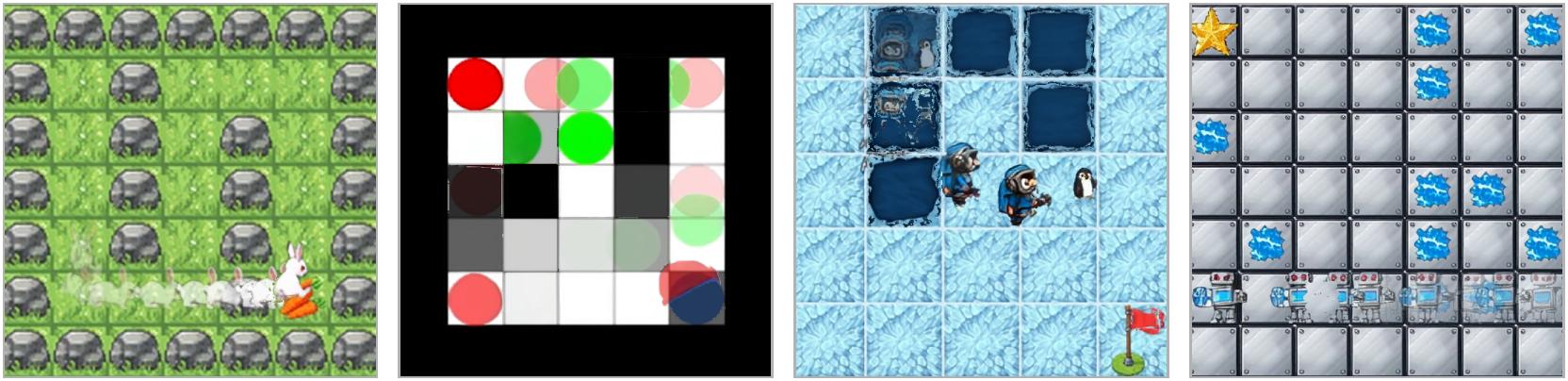}
\caption{\textbf{Constraint violations --- VR-Bench.} The agent enters a forbidden cell
(lake or trap) in procedurally generated mazes and trapfields. Each panel uses a different
texture to illustrate that the failure mode is consistent across visual appearances.}\end{figure}

\begin{figure}[H]\centering
\includegraphics[width=0.88\textwidth]{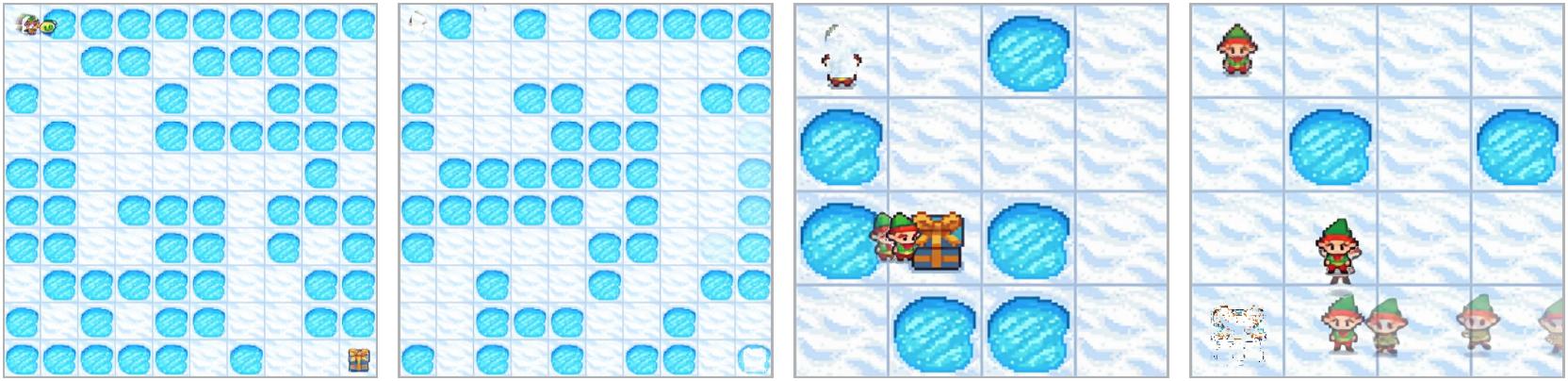}
\caption{\textbf{Degenerate failures.} The model produces a video with little or no
meaningful agent motion. The ghost trail collapses to a single position, indicating that
the elf remains static throughout generation.}\end{figure}

}

\begin{figure}[H]\centering
\includegraphics[width=0.88\textwidth]{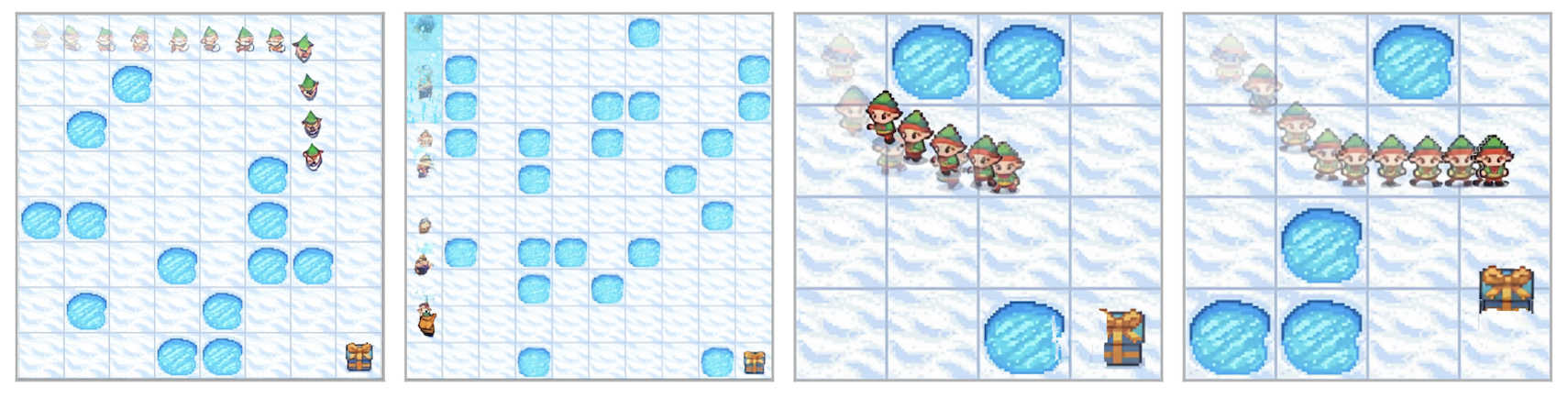}
\caption{\textbf{Horizon-limited failures.} The generated trajectory respects all
constraints but fails to reach the goal within the 81-frame budget. \emph{Valid stall}:
the agent follows a correct prefix but stops moving before reaching the goal.
\emph{Wrong route}: the agent takes a legal but sub-optimal path and runs out of frames.
}\end{figure}